\documentclass[lettersize,journal]{IEEEtran}
\usepackage{graphicx}
\usepackage[numbers]{natbib}
\usepackage{amsthm}
\newtheorem{remark}{Remark}
\usepackage{xcolor}
\usepackage{amsmath,amsfonts}
\usepackage{algorithmic}
\usepackage{algorithm}
\usepackage{array}
\usepackage{tikz}
\usepackage[caption=false,font=normalsize,labelfont=sf,textfont=sf]{subfig}
\usepackage{textcomp}
\usepackage{stfloats}
\usepackage{url}
\usepackage[compatibility=false]{caption}
\usepackage{balance}
\usepackage{blindtext}
\usepackage{tikz}
\usepackage{caption}
\usepackage{amsmath}
\usepackage{enumitem}
\setlist[itemize]{nosep}  
\usepackage{verbatim}
\usepackage{amssymb}
\usepackage{booktabs}
\usepackage[skip=2pt]{caption}
\usetikzlibrary{positioning, shapes, arrows}
\usepackage{verbatim}
\usepackage{graphicx}
\usepackage{cite}
\begin{document}

\title{Image Categorization and Search via a GAT Autoencoder and Representative Models}

\author{%
    Duygu~Sap$^{*}$,\ %
    Martin~Lotz,\ %
    and~Connor~Mattinson%
    \thanks{Duygu Sap$^{*}$ is with CAMaCS, Mathematics Institute, University of Warwick, Coventry, United Kingdom 
    (e-mail: duygu.sap@warwick.ac.uk).}%
    \thanks{Martin Lotz is with the Mathematics Institute, University of Warwick, Coventry, United Kingdom 
    (e-mail: martin.lotz@warwick.ac.uk).}%
    \thanks{Connor Mattinson is with TRUSS, London, United Kingdom 
    (e-mail: connor@trussarchive.com).}%
    \thanks{*$\,$Corresponding author.}%
}

\maketitle





\begin{abstract}
We propose a method for image categorization and retrieval that leverages graphs and a graph attention network (GAT)-based autoencoder. Our approach is representative-centric, that is, we execute the categorization and retrieval process via the representative models we construct for the images and image categories. We utilize a graph where nodes represent images (or their representatives) and edges capture similarity relationships. GAT highlights important features and relationships between images, enabling the autoencoder to construct context-aware latent representations that capture the key features of each image relative to its neighbors. We obtain category representatives from these embeddings and categorize a query image by comparing its representative to the category representatives. We then retrieve the most similar image to the query image within its identified category. We demonstrate the effectiveness of our representative-centric approach through experiments with both the GAT autoencoders and standard feature-based techniques.  
\end{abstract}

\begin{IEEEkeywords}
Image categorization, Image comparison, GAT, Graphs, Autoencoders, Representative model
\end{IEEEkeywords}

\section{Introduction}
\noindent The comparison and categorization of images has been one of the fundamental challenges addressed by deep learning methods. Convolutional Neural Networks (CNNs) have been a primary deep learning method used in tackling tasks that require image identification or classification due to their ability to automatically extract meaningful features from images. By using convolutional layers, CNNs can detect patterns such as edges, textures, and shapes, which are essential for tasks such as object recognition, facial verification, and anomaly detection \citep{lecun2015deep}. A key success of CNNs was AlexNet, which outperformed previous methods in the ImageNet competition, significantly reducing error rates \citep{krizhevsky2012imagenet}. Building on this success, ResNet \citep{he2016} introduced deep residual learning, which enabled the training of significantly deeper networks and led to improvements in both accuracy and efficiency. Despite their successful applications, CNNs are limited by their reliance on grid-based representations and local receptive fields, which restrict their ability to capture complex relationships between distant regions within an image.\\
\indent To address these limitations of CNNs, Graph Neural Networks (GNNs) have emerged as an alternative approach that is capable of modeling non-Euclidean relationships in data \citep{gori2005}. \citep{zhang2018deeplearning},\citep{wu2019comprehensive}, \citep{chami2020mlongraphs} offer surveys on GNNs. While \citep{wu2019comprehensive} categorize GNNs into four groups: recurrent graph neural networks, convolutional graph neural networks, graph autoencoders, and spatial-temporal graph neural networks, \citep{zhang2018deeplearning} offer a systematic overview of different graph deep learning methods and \citep{chami2020mlongraphs} propose a graph encoder decoder model to unify network embedding and graph neural network models. \\
\indent GNN architectures such as Graph Convolutional Networks (GCN) \citep{kipf2017semi}, Graph Attention Networks (GAT) \citep{velickovic2018graph}, and GraphSAGE \citep{hamilton2017inductive} have demonstrated good performance in tasks requiring relational reasoning. By representing images as graphs and leveraging message passing techniques, GNNs can capture both local and global dependencies more effectively than CNNs.  In a GNN framework, an iterative process propagates node states until equilibrium is achieved, and then a neural network produces an output for each node based on its state. More precisely, in GNNs, information is passed iteratively between nodes in a process known as message passing. This message passing process continues until the node states, which is the information each node holds, converge or reach equilibrium so that a stable representation of the graph or image can be obtained. After this process, a neural network is used to process the final states of the nodes and generate an output, which is typically a classification or prediction. GNN framework can further be enhanced with the use of attention mechanisms. Attention mechanisms allow models to focus on the most relevant parts of the input, which is particularly useful while dealing with inputs of varying sizes or complex structures. For example, attention can help the model focus on the relevant parts of an image (such as an object in the image) while ignoring irrelevant parts (such as the background). A key attention mechanism is self-attention (or intra-attention), which computes a representation of a sequence by attending to its own elements, capturing long-range dependencies and contextual relationships within the data. This mechanism has been successfully applied in natural language processing tasks, particularly through the transformer architecture \citep{vaswani2017attention}, which revolutionized machine translation and other sequence-based tasks. GATs extend this self-attention mechanism to graph-structured data. They  were built on this transformer architecture. In GATs, the attention mechanism is used to compute the hidden representations of each node by attending to its neighbors, dynamically assigning different attention weights based on their relevance. This approach offers several advantages: it is computationally efficient due to its parallelizability across node-neighbor pairs, it can handle nodes with varying degrees by assigning different weights to neighbors, and it is well-suited for inductive learning, allowing the model to generalize to completely unseen graphs \citep{romero2018}. Thus, GATs can be used effectively to draw attention to the most relevant features of the images, and improve the model's ability to make accurate comparisons and classifications.\\
\indent In this research, we build a graph where nodes represent images and edges capture similarity relationships, and use a GAT-based autoencoder (GAT-AE) to construct representative models for a given set of image listings based on the significant features and connections within the images. Then, we categorize the representative models, thus, the image listings, to establish predefined categories. We demonstrate applications of our method in identifying the categories of  query images by comparing their representatives against these pre-categorized representative models, and retrieving the most similar image within the categories identified for the query images.\\
\indent While our approach is based on GAT autoencoders operating over graph-structured image data, other attention-based models in computer vision follow a different paradigm. For example, the Contrastive Language–Image Pretraining (CLIP) model \citep{radford2021learning} aligns image and text embeddings using dual transformer-based encoders, and FashionCLIP \citep{chambon2022fashionclip} fine-tunes CLIP for fashion-specific image–text pairs. These models rely on transformer-style attention over sequences, such as tokens or image patches, rather than graph-structured attention, and like standard GATs, they were not originally designed for autoencoder frameworks. In this research, we adapt GATs into an autoencoder framework to explicitly leverage graph-structured attention over image features, where node attributes are extracted using a backbone encoder.  We can use the node features coming from ResNet, CLIP’s vision encoder, or any other CNN/transformer-based model in our pipeline. Our approach offers advantages particularly in scenarios where spatial relationships or structural constraints play a key role.\\
\indent The rest of the manuscript is outlined as follows: in Section~\ref{sec:preliminaries}, we present the basic concepts of GNNs, GCNs, graph autoencoders (GAEs), GATs; in Section~\ref{sec:methodology}, we describe our methodology and algorithmic frameworks; in Section~\ref{sec:applications}, we provide applications of our representative-centric GAT-AE based method along with the applications of our representative-centric method using some conventional methods; and in Section~\ref{sec:conclusion}, we provide a conclusion with remarks and comments on our results. 

\section{Preliminaries}\label{sec:preliminaries}
\noindent In this section, we present the basic concepts of Graph Neural Networks (GNN), Graph Convolutional Neural Networks (GCN), Graph Attention Networks (GAT), and Graph Autoencoders (GAE).
\subsection{Graph Neural Networks (GNN)}
\noindent Graph neural networks operate primarily with the following components:
\begin{itemize}
    \item $G =(V,E)$: a graph with the set of nodes $V$ and edges $E$,
    \item $X$: node feature vectors,
    \item $A$: an adjacency matrix that stores the connections between nodes,
    \item $M$: a learnable function that computes the message to be sent.
    \item $Z$: node embeddings obtained after the encoder stage, representing nodes in a lower-dimensional latent space.
\end{itemize}
The key stage of GNNs is the pairwise message-passing process, which iteratively updates each node's representation by exchanging information with its neighbors. During this process, each node sends a message to its neighbors which are then aggregated to produce an updated feature representation for each node. The message passed from node $i$ to node $j$ is typically a function of the features of nodes $i$ and $j$, and the edge connecting them.\\ The general state update for node $i$ can be listed as follows:
\begin{equation*}
    h_i^{(n+1)} = \sigma\left( \sum_{j \in \mathcal{N}_i} M\left(h_i^{(n)}, h_j^{(n)}, e_{ij}\right) \right),
\end{equation*}
where $\mathcal{N}_i$ is the set of neighbors of node $i$, $h_i^{(n)}$ is the feature vector of node $i$ at layer $n$, $e_{ij}$ is the feature vector of the edge between $i$ and $j$, and $\sigma(\cdot)$ is a nonlinear activation function.
This process is typically repeated for several iterations, allowing each node to gather information from increasingly distant nodes in the graph. The final output of the GNN can be obtained by feeding the updated feature representations of all nodes through a fully connected layer or another appropriate output layer \citep{rice2023gnnlecture}.
\begin{figure}[H]
    \centering
    \includegraphics[width=\linewidth]{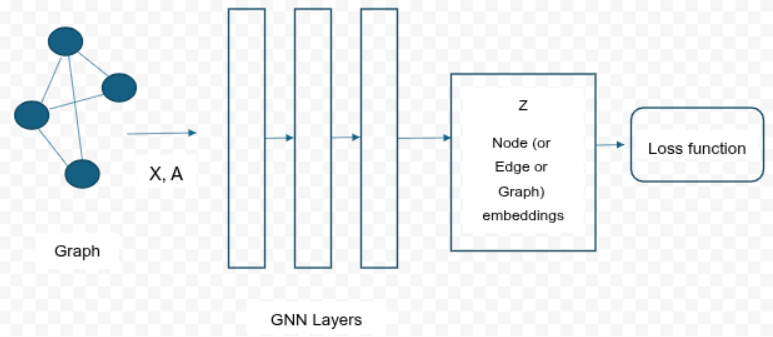}
    \caption{A general GNN pipeline}
    \label{fig:GNN_pipeline}
\end{figure}
\subsubsection{Graph Convolutional Networks (GCN)}
\noindent Graph Convolutional Networks (GCN) are a special type of GNN where the message-passing process is implemented using a fixed neighborhood aggregation based on the graph’s normalized adjacency matrix. GCN has the following additional components\citep{kipf2017semi}:
\begin{itemize}
    \item $\tilde{A} = A + I$: adjacency matrix of the undirected graph $G$ with added self-connections where $I$ is the identity matrix,
    \item $\tilde{D}$: degree matrix of $\tilde{A}$, where $\tilde{D}_{ii} = \sum_{j} \tilde{A}_{ij}$ and $\tilde{D}_{ij}=0$ for all $i\neq j$. 
    \item $W^{(l)}$: a layer-specific trainable weight matrix.
\end{itemize}
A single GCN layer updates node features as follows:
\begin{equation*}
    h^{(l+1)} = \sigma\left( \tilde{D}^{-\frac{1}{2}} \ \tilde{A} \ \tilde{D}^{-\frac{1}{2}} \ h^{(l)} W^{(l)} \right),
\end{equation*}
where $h^{(0)} := X$. This symmetric normalization prevents high-degree nodes from dominating the aggregation and ensures numerical stability.

\subsubsection{Graph Attention Networks (GAT)}
\noindent Graph Attention Network (GAT) \citep{velickovic2018graph} incorporates the attention mechanism into the propagation step of GNN. It computes the hidden states of each node by attending to its neighbors, following a self-attention strategy. The hidden state of node $v$ can be obtained by \citep{zhou2020gnnreview}:
\begin{equation}
    h^{t+1}_v = \sigma \left( \sum_{u \in \mathcal{N}_v} \alpha_{vu} W h^t_u \right),
\end{equation}
where
\begin{equation}
    \alpha_{vu} = \frac{\exp \left( \mathrm{LeakyReLU} \left( \mathbf{w}^\top [W h_v \, \| \, W h_u] \right) \right)}
    {\sum_{k \in \mathcal{N}_v} \exp \left( \mathrm{LeakyReLU} \left( \mathbf{w}^\top [W h_v \, \| \, W h_k] \right) \right)}.
    \label{eq:gat_attention}
\end{equation}
Here, $W$ is the weight matrix associated with the linear transformation applied to each node, and $\mathbf{w}$ is the weight vector of a single-layer multi-layer perceptron (MLP). In this context, the MLP is simply a linear projection followed by a nonlinearity, rather than a deep network with multiple hidden layers.  \\
In addition, GAT can utilize the multi-head attention mechanism proposed by \citep{vaswani2017attention} to stabilize the learning process. In this setup, $K$ independent attention heads are applied in parallel to compute node representations, and their outputs are either concatenated as in \eqref{eq:multi-heads} or averaged as in \eqref{eq:aver-heads} to form the final feature vector.
 \begin{equation}
    h^{t+1}_v = \big\|_{k=1}^K \sigma \left( \sum_{u \in \mathcal{N}_v} \alpha^{k}_{vu} W^{k} h^t_u \right).
    \label{eq:multi-heads}
\end{equation}
\begin{equation}
    h^{t+1}_v = \sigma \left( \frac{1}{K} \sum_{k=1}^K \sum_{u \in \mathcal{N}_v} \alpha^{k}_{vu} W^{k} h^t_u \right).
    \label{eq:aver-heads}
\end{equation}
Here, $\big\|_{k=1}^K$ denotes vector concatenation over the $K$ attention heads, and $\alpha^{k}_{ij}$ is the normalized attention coefficient computed by the $k$-th attention head \citep{zhou2020gnnreview}. 
\subsection{Graph Autoencoders (GAE)}
\begin{figure}[H]
    \centering
    \includegraphics[width=\linewidth]{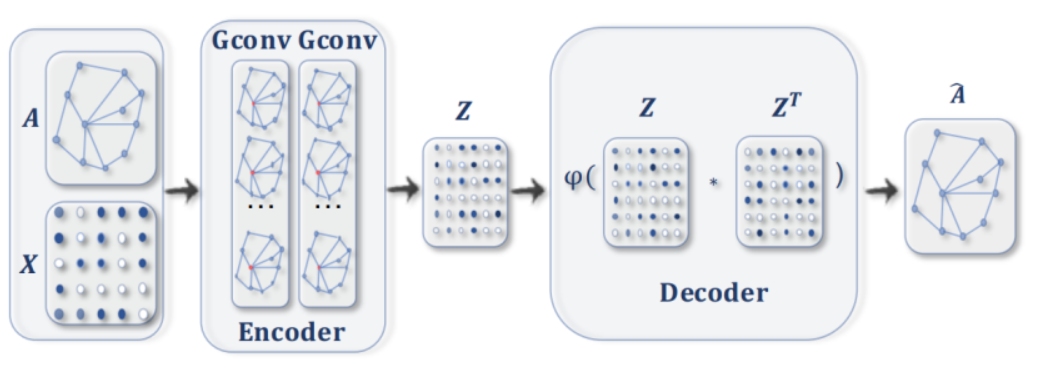}
    \caption{ GAE for network embedding (Image courtesy of \citep{rice2023gnnlecture})}
    \label{fig:gae}
\end{figure}
\noindent Graph autoencoders are unsupervised learning models that operate without the need for labeled data. They employ an encoder–decoder architecture in which nodes or entire graphs are mapped into a latent vector space and then reconstructed from the encoded representations. In the context of network embedding, GAEs can be used to learn latent node representations by reconstructing graph structural information such as the graph adjacency matrix \citep{rice2023gnnlecture}. In the GAE architecture illustrated in Figure~\ref{fig:gae}, the encoder employs graph convolutional layers to generate a network embedding for each node. The decoder then uses these embeddings to compute pairwise distances. After applying a non-linear activation function, the decoder reconstructs the graph adjacency matrix. The network is trained by minimizing the difference between the actual adjacency matrix and the reconstructed adjacency matrix.\\
In this research, we adopt a graph attention autoencoder (GAT-AE), which is a special type of GAE where the encoder is a GAT. Unlike standard graph autoencoders, which often employ GCN-based encoders with fixed neighborhood weights, our approach uses attention-based message passing to learn adaptive, context-dependent neighbor importance scores. This design is particularly suitable for our setting, where node features are extracted from images and neighbor relevance can vary significantly across the graph.
\section{Methodology}\label{sec:methodology}
\noindent Let $\mathcal{G}$ be a directory that consists of listings of images. We denote each listing by $\mathcal{L}_i$ and the representative of listing $\mathcal{L}_i$ by $\mathcal{R}_i$. $\mathcal{R}_i$ can be an abstract model that is not necessarily present in the listing. It serves as a proxy for the image, conceptually similar to the abstract proxies defined for CAD models in \citep{SAP2019256,SAP2025101365}. \\
\indent The categories of images may be predefined or we may define them by grouping $\{\mathcal{L}_i\}$ utilizing their representatives $\{\mathcal{R}_i\}$ and a similarity threshold value. \\\indent When the categories are predefined,  we denote by $\mathcal{G}$  the main category,  and let $\mathcal{L}_i$ denote a subcategory of $\mathcal{G}$, thus, $\mathcal{R}_i$ becomes the representative of the subcategory $\mathcal{L}_i$. For brevity, we simply refer to the subcategories $\mathcal{L}_i$  as "categories" in this case.
\begin{figure}[H]
    \centering
    \includegraphics[width=\linewidth]{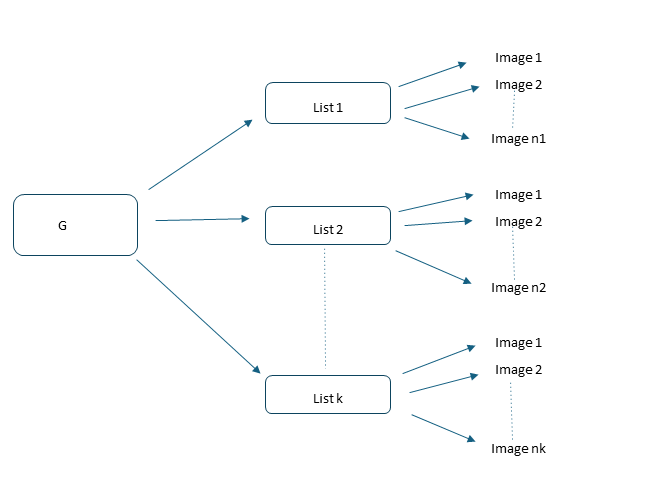}
    \caption{Image Data. Here $G$ denotes a category, $k$ denotes the number of subcategories of $G$ and $n_k$ denotes the number images in the subcategory denoted by $k$}
    \label{fig:fashion-data}
\end{figure}
\noindent Our methodology mainly consists of two stages: 
\begin{enumerate}
    \item Representative Model Construction, 
    \item Graph Construction and Model Categorization.
    \end{enumerate}
We follow two slightly distinct approaches to construct the representative models of the listings or categories. Each approach utilizes a GAT autoencoder (GAT-AE) that enables deriving context-aware latent representations of the images.
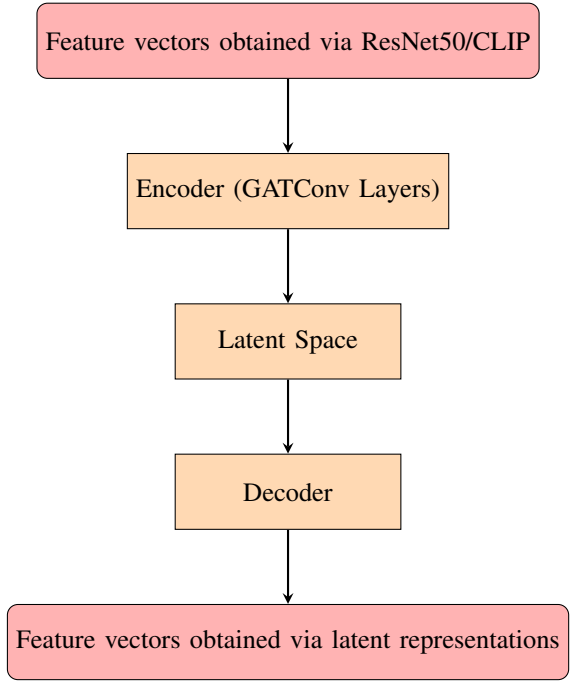
\begin{figure}[ht]
    \centering
\begin{tikzpicture}[node distance=2cm]

\tikzstyle{startstop} = [rectangle, rounded corners, minimum width=3cm, minimum height=1cm,text centered, draw=black, fill=red!30]
\tikzstyle{process} = [rectangle, minimum width=3cm, minimum height=1cm, text centered, draw=black, fill=orange!30]
\tikzstyle{decision} = [diamond, minimum width=3cm, minimum height=1cm, text centered, draw=black, fill=yellow!30]
\tikzstyle{arrow} = [thick,->,>=stealth]

\node (start) [startstop] {Feature vectors obtained via ResNet50/CLIP};
\node (process1) [process, below of=start] {Encoder (GATConv Layers) };
\node (process2) [process, below of=process1] {Latent Space};
\node (process3) [process, below of=process2] {Decoder};
\node (stop) [startstop, below of=process3] {Feature vectors obtained via latent representations};

\draw [arrow] (start) -- (process1);
\draw [arrow] (process1) -- (process2);
\draw [arrow] (process2) -- (process3);
\draw [arrow] (process3) -- (stop);

\end{tikzpicture}

 \caption{The GAT Autoencoder (GAT-AE)}
    \label{fig:gat-auto} 
\end{figure}

\subsection{Approach I }\label{sec:approach-I}

\subsubsection{Representative Model and Graph Construction}\label{sec:representative_model_I}
We employ the GAT-AE whose structure is illustrated in Figure~\ref{fig:gat-auto} to determine the essential features within each image $m_{ij}$ in a listing $\mathcal{L}_i$, and then encode these features along with the relevant connections into an image, $m_{ij}^r$, which can serve as  the context-aware latent vector (or the representative) of $m_{ij}$.
We assess the centrality of each image representative  by summing their quantified similarity to others using a cosine similarity matrix, and identify the image with the representative $m_{ij}^r$  that has the highest total similarity to all other image representatives as the representative $\mathcal{R}_i$ of the listing. 
\begin{remark}
    One can calculate how exceptional the representative image is in comparison to others by calculating the z-score. It is also possible to show a ranked list of the top $k$ most central images in the set to illustrate other images that are also highly representative of the group.
\end{remark}

\subsubsection{Model Categorization}\label{sec:I-model_category}
\noindent As mentioned earlier, we may or may not have the categories predefined, so we provide a strategy to use in each case below. 

\begin{itemize}
    \item \textbf{Case 1.} The categories are not predefined. In this scenario, after deriving a representative model $\mathcal{R}_i$ for each listing $\mathcal{L}_i$, we compare the listings $\{\mathcal{L}_j\}$ via their representatives $\{\mathcal{R}_j\}$ to define the categories. We determine the categories by following one of the two approaches.
    \begin{enumerate}
        \item Categorize the representatives (thus, the listings) by collecting those with similarity measures above a predefined threshold in the same category. 
        \item Categorize the representatives (thus, the listings)  via some well-known clustering algorithms such as K-Means or DBSCAN. 
        \end{enumerate}
        Then, we identify the most central listing representative of each category as the representative of the category.
Once the images are categorized, we determine the category of the query image by comparing its context-aware latent derived as described in Section~\ref{sec:representative_model_I} via the same GAT-AE with the representatives of the categories.
\item \textbf{Case 2.} The categories are predefined. In this scenario, we can proceed directly to determine the category of the query image, as described in Case 1. In this case, we refer to the listings as categories and denote by $\mathcal{R}_i$ the representative of the $i^{th}$ category as mentioned earlier.
\end{itemize}
\begin{remark}
    The first level of our graph-centric approach captures the similarities between the images within each listing. When the categories are not predefined, the second level of our graph-centric approach first compares the representatives of different listings to identify the categories, and then uses the distinctions between these representatives to determine the category of the query image. On the other hand, when the categories are predefined, the second level of our graph-centric approach focuses only on comparing the representatives with the query image, relying solely on the distinctions between the representatives. 
\end{remark}

\subsection{Approach II}
This approach is also representative-centric and utilizes the GAT-AE that we introduced to enable content-based image retrieval using a graph-aware latent space. However, unlike the first approach, we use a pretrained ResNet model and derive the category representatives by averaging the image representatives rather than selecting the most central image. \\
\indent We assume that the categories are predefined. Precisely, there is a main category with predefined subcategories. 
\subsubsection{Representative Model and Graph Construction}\label{sec:approach_ii_rep_construct}
First, we extract high-level visual features from all dataset images using a pre-trained ResNet-50 model, and then use these features to build a k-NN graph to capture the similarity relationships between images. The nodes of this k-NN graph are the ResNet feature vectors. Second, we train a GAT-AE on this graph to learn the \textit{context-aware latent vectors}, $\{m_{ij}^r\}$, which are compact representations that reflect both image content and neighborhood structure. For each category, we compute a representative vector by averaging $\{m_{ij}^r\}_{j=1}^{n_i}$, where $n_i$ denotes the number of images in the category denoted by $\mathcal{L}_i$. 
\subsubsection{Model Categorization}
To categorize the query image, we first extract its ResNet features and pass them through the GAT-AE with self-loops only. This yields a context-free latent representation for the query image. We then compare this resulting latent representation to category representatives and identify the category of the query image.\\ 
To retrieve the image that is most similar to the query image within the determined category $\mathcal{L}_i$, we insert the ResNet feature vector of the query image into the k-NN graph constructed with the ResNet features vectors of the images in $\mathcal{L}_i$. We then run the GAT-AE to generate a \textit{context-aware latent vector} of the query image. After that, we compare this latent to the context-aware latents $\{m_{ij}^r\}_{j=1}^{n_i}$ of the images in $\mathcal{L}_i$.
\begin{remark}
    Note that our methodology provides a flexible GAT-AE model, that is, the same model can encode new, unseen images as long as we connect them to existing nodes.
\end{remark}
\section{Applications}\label{sec:applications}
In this section, we present some applications of our methodology described in Section~\ref{sec:methodology} along with some applications of our representative-centric approach to more conventional methods.
\subsection{Applications of Approach I}\label{sec:apps-I}
\noindent In these applications, we focus on finding a representative model for a given listing and executing the image search and categorization for query images. We also illustrate how to categorize listings as described in Section~\ref{sec:approach-I} for the case where the image categories are unknown. We demonstrate the application of our GAT-AE–based approach and present alternative methods for constructing representatives, and performing image search and categorization. \\
\indent In the experiments below, we assume that the main category is predefined; precisely, $\mathcal{G}$ is defined as \textit{clothes}, and we have three subcategories. We use the Fashion Product Images Dataset available at \citep{fashion_kaggle}.
Experiment~\ref{exp:Approach-I-1} applies our GAT-AE–based method on CLIP features to construct representatives and categorize query images, while Experiment~\ref{exp:Approach-I-3} uses CLIP features directly without the GAT-AE, and Experiment~\ref{exp:Approach-I-2} employs the \textit{histogram of oriented gradients} method. In Experiment~\ref{exp:Approach-I-2}, we also consider the case where the subcategories are not predefined and illustrate how categorization can be executed using image listings selected from DeepFashion's In-Shop Clothes Retrieval Bench available at \citep{fashion_deepfashion_inshop}.

\begin{figure}[H]
    \centering
    \resizebox{\columnwidth}{!}{
        \begin{tabular}{|c|c|c|c|c|c|c|c|}
            \hline
            \includegraphics[width=0.1\textwidth]{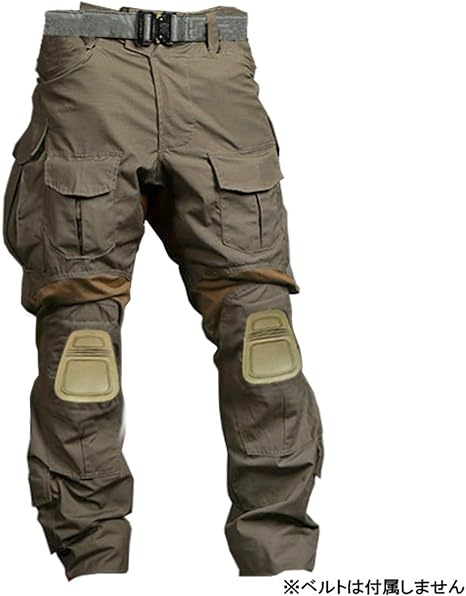} &
            \includegraphics[width=0.1\textwidth]{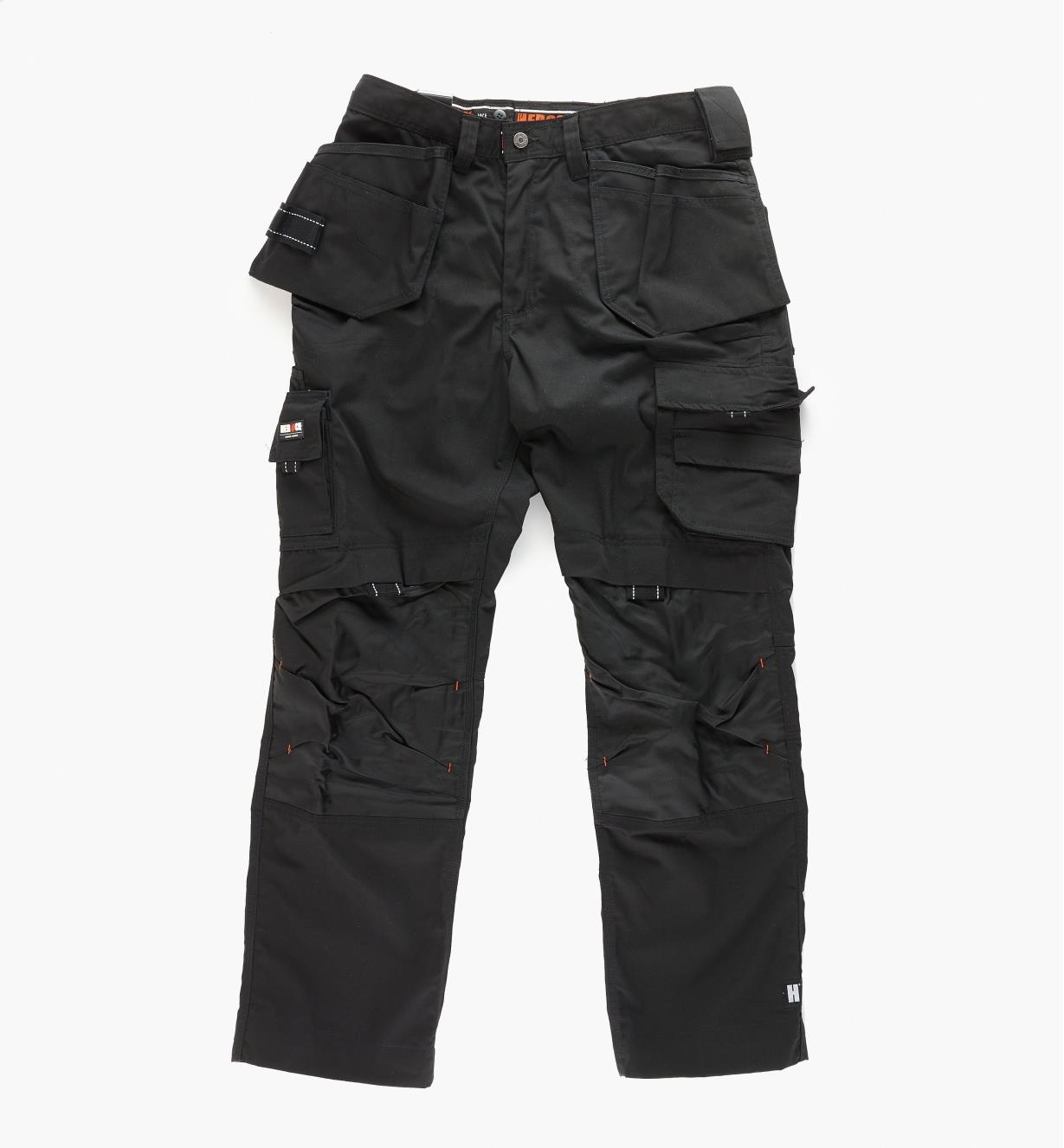} &
            \includegraphics[width=0.1\textwidth]{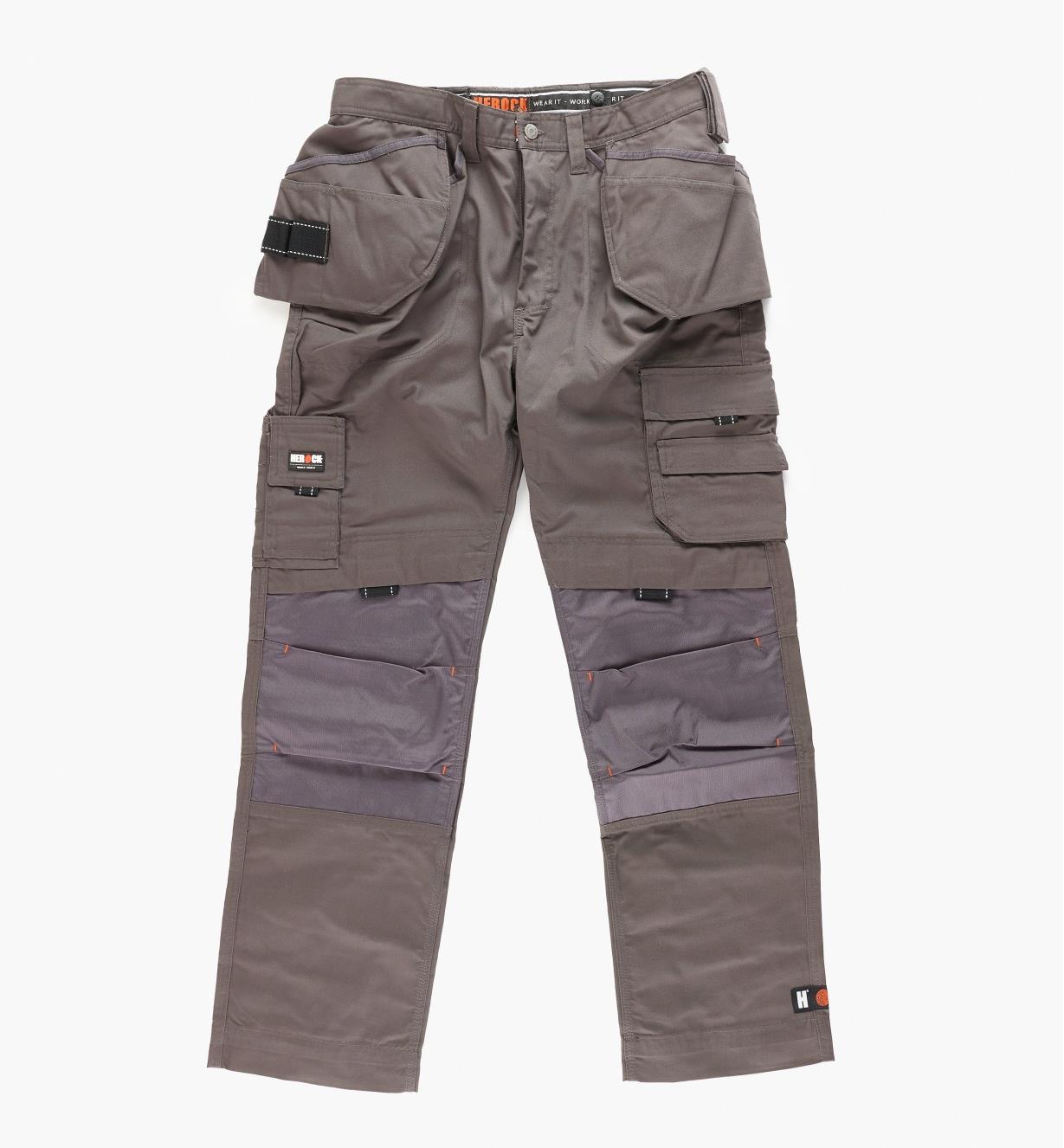} &
            \includegraphics[width=0.1\textwidth]{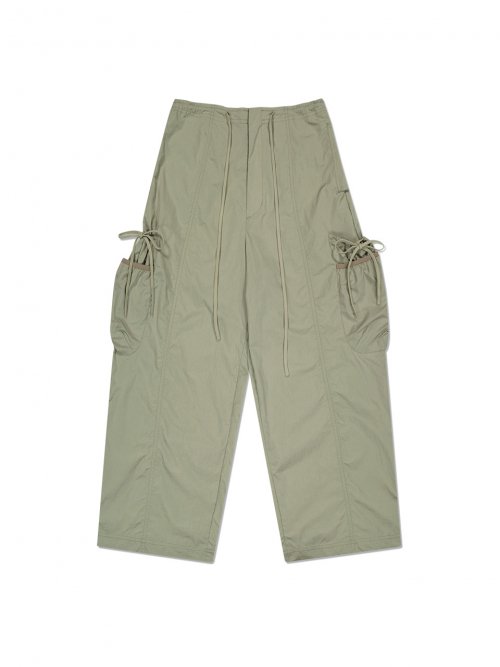} &
            \includegraphics[width=0.1\textwidth]{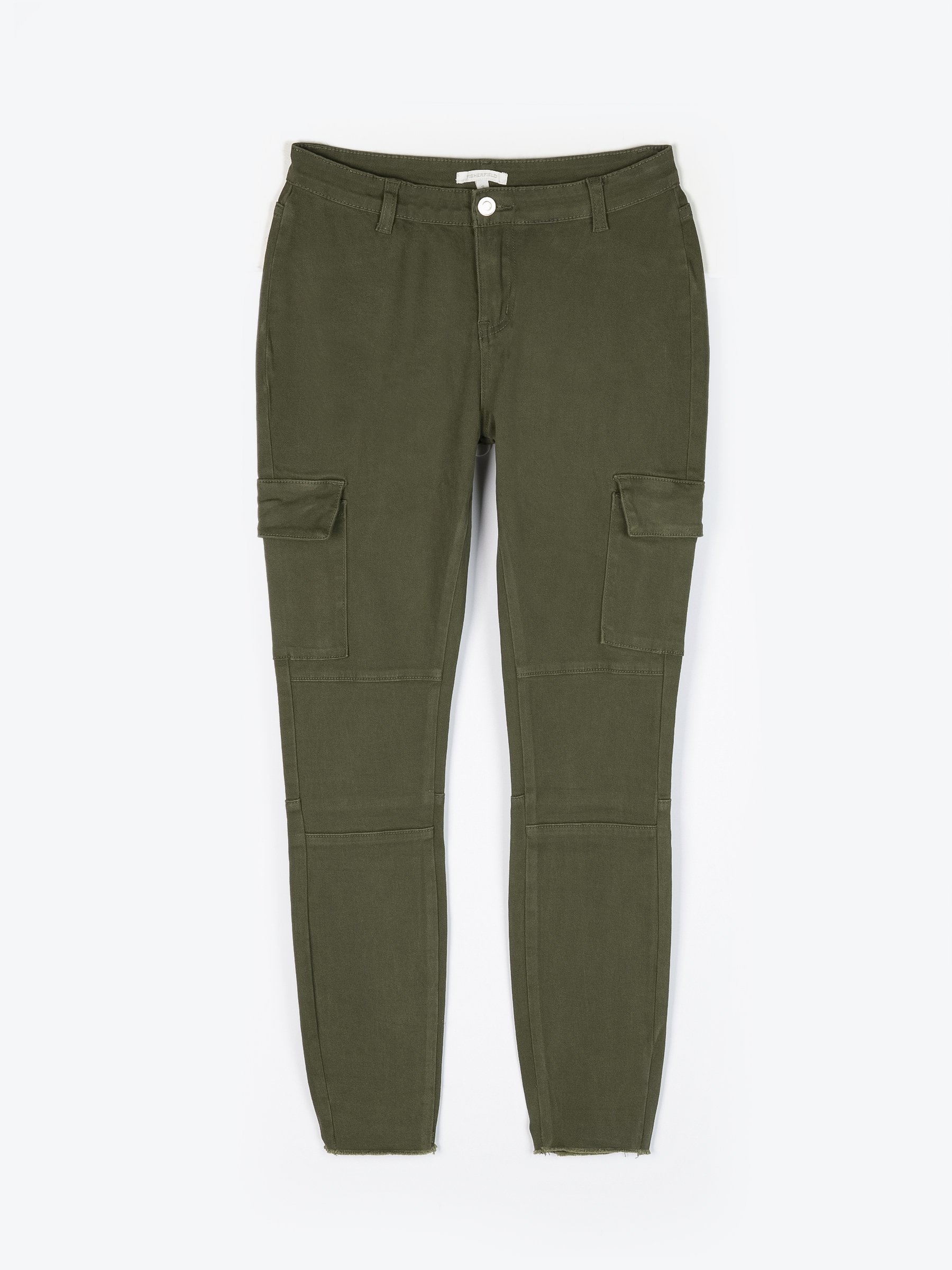} &
            \includegraphics[width=0.1\textwidth]{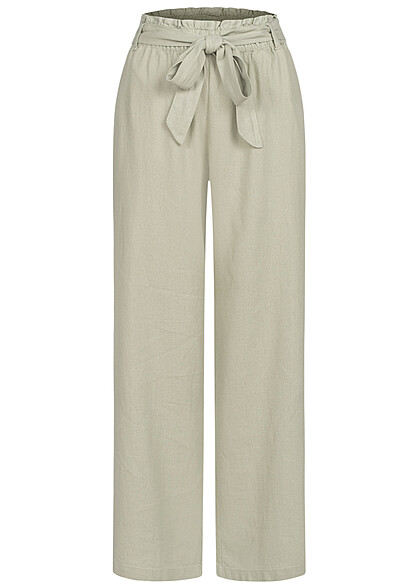} &
            \includegraphics[width=0.1\textwidth]{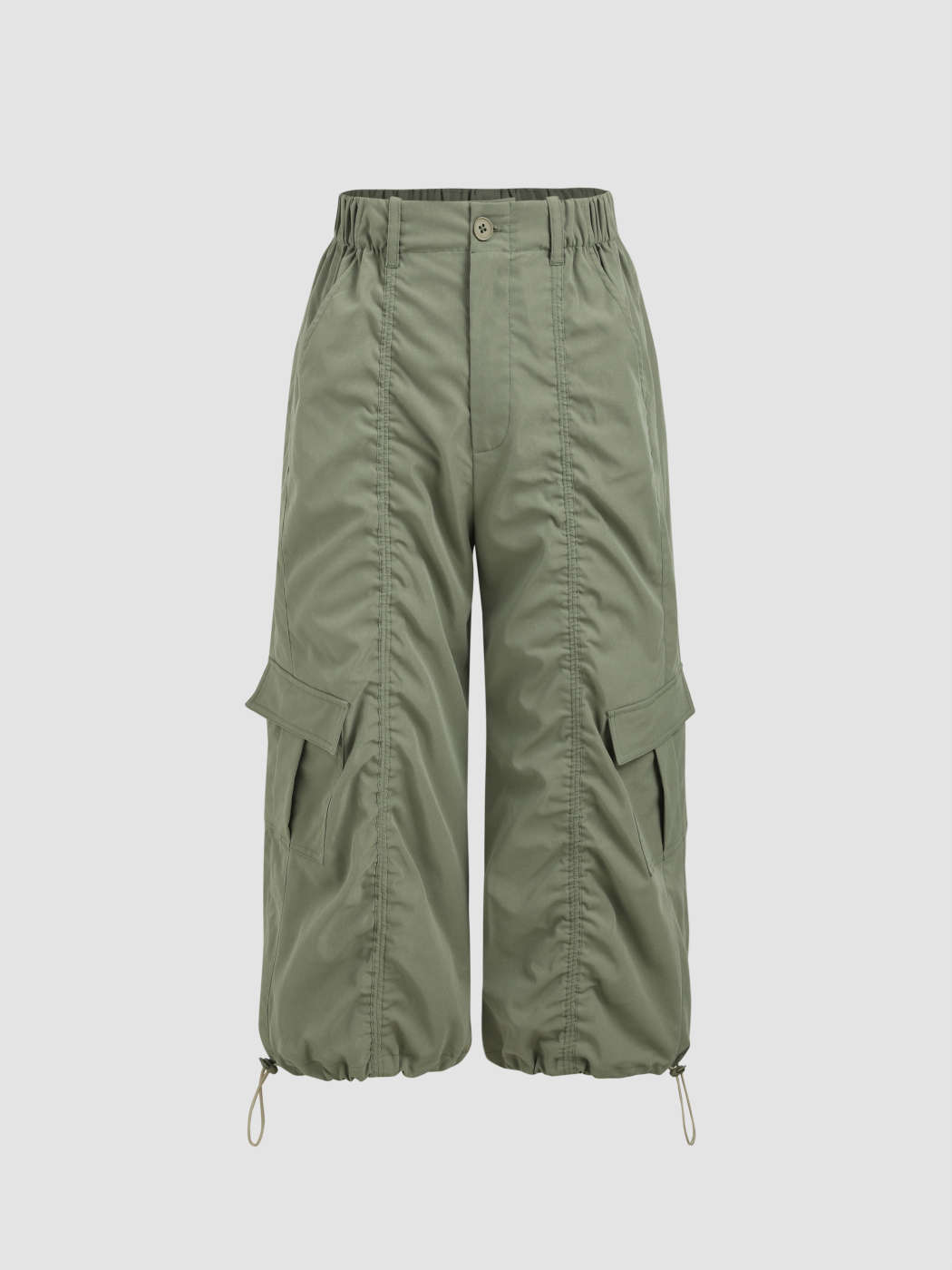} &
            \includegraphics[width=0.1\textwidth]{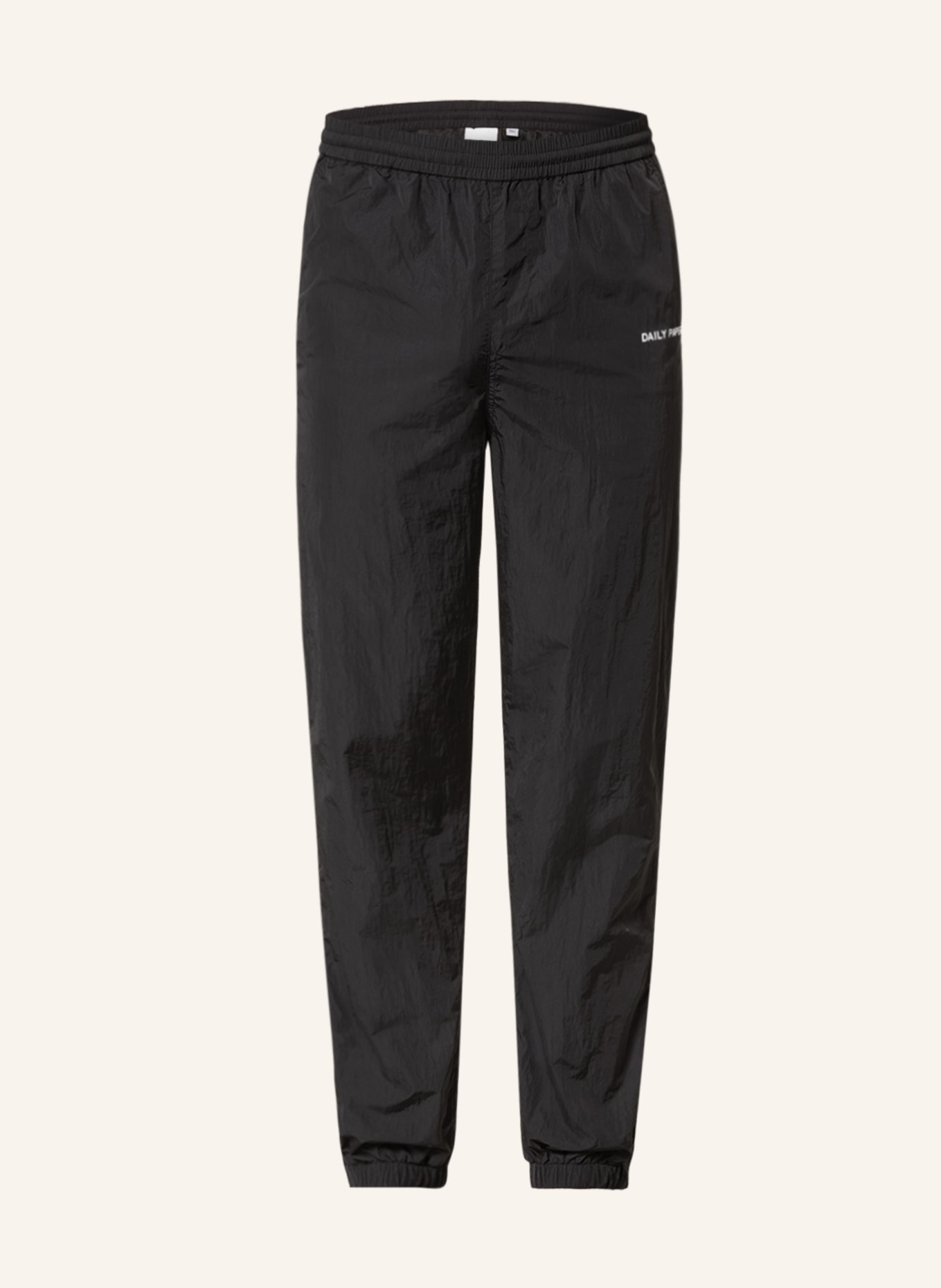} \\
            \hline
            \includegraphics[width=0.1\textwidth]{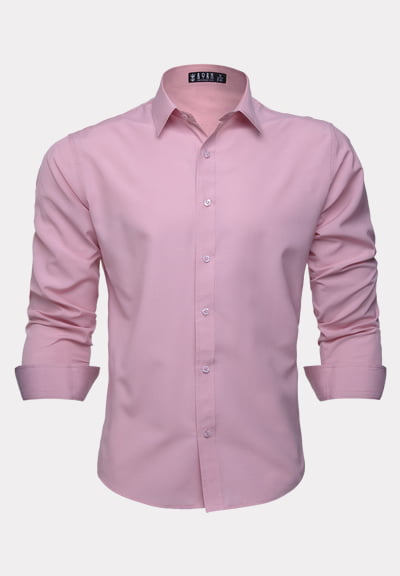} &
            \includegraphics[width=0.1\textwidth]{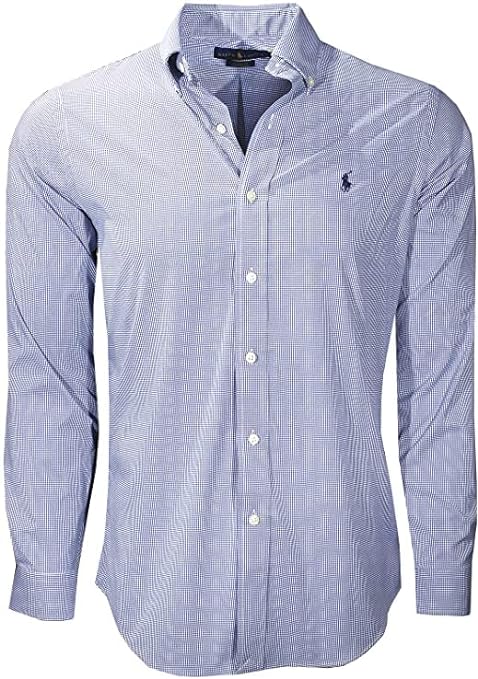} &
            \includegraphics[width=0.1\textwidth]{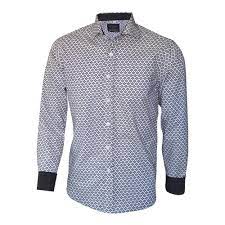} &
            \includegraphics[width=0.1\textwidth]{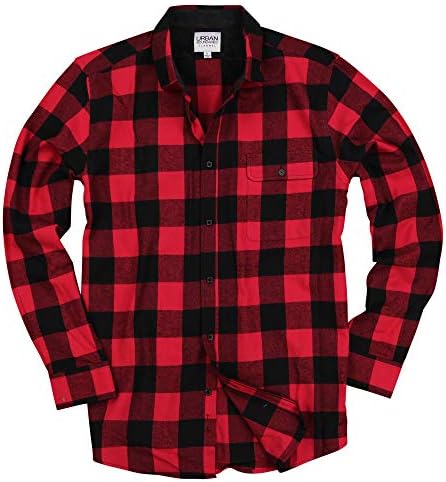} &
            \includegraphics[width=0.1\textwidth]{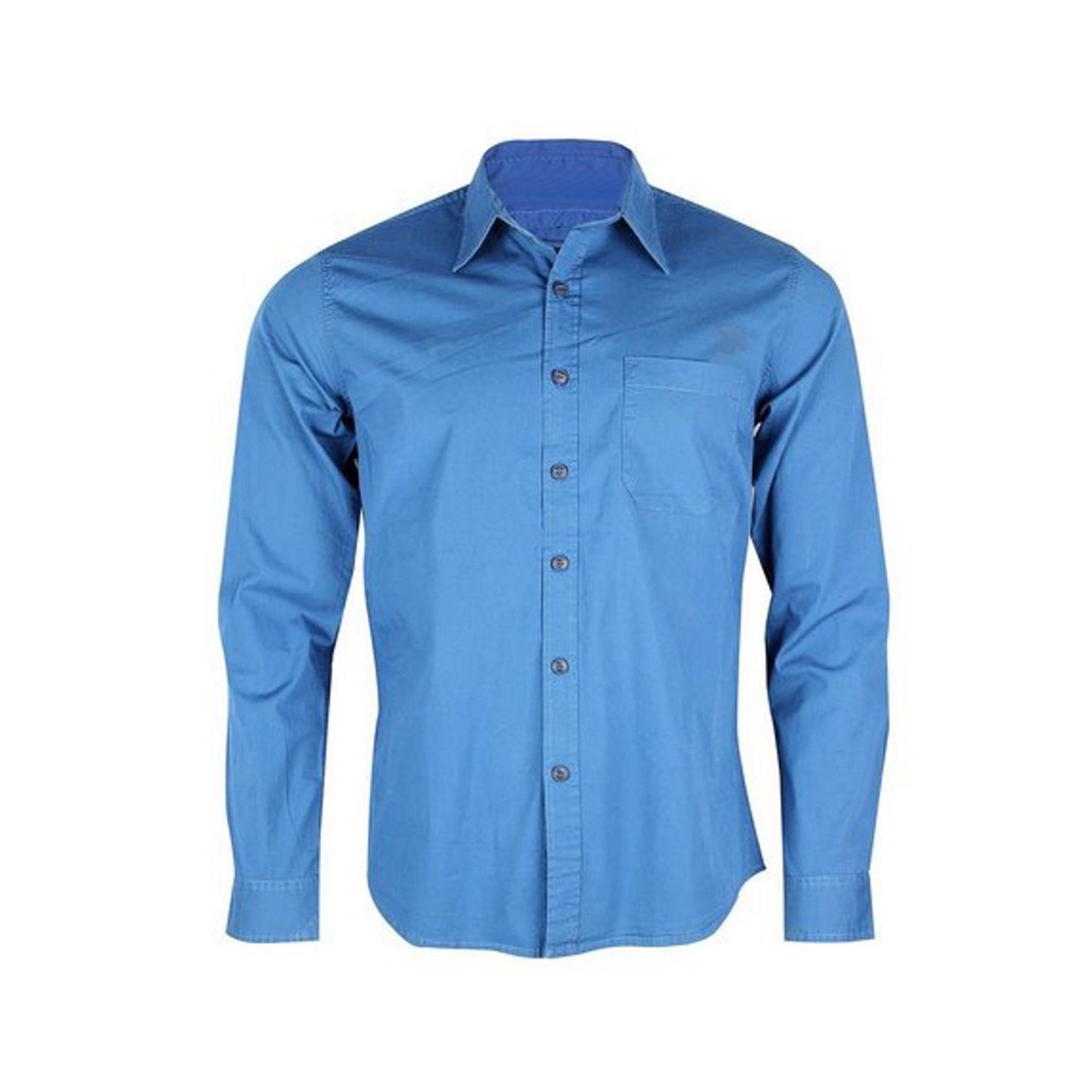} &
            \includegraphics[width=0.1\textwidth]{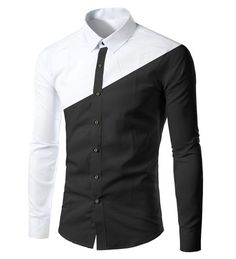} &
            \includegraphics[width=0.1\textwidth]{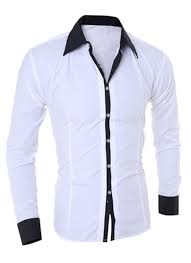} &
            \includegraphics[width=0.1\textwidth]{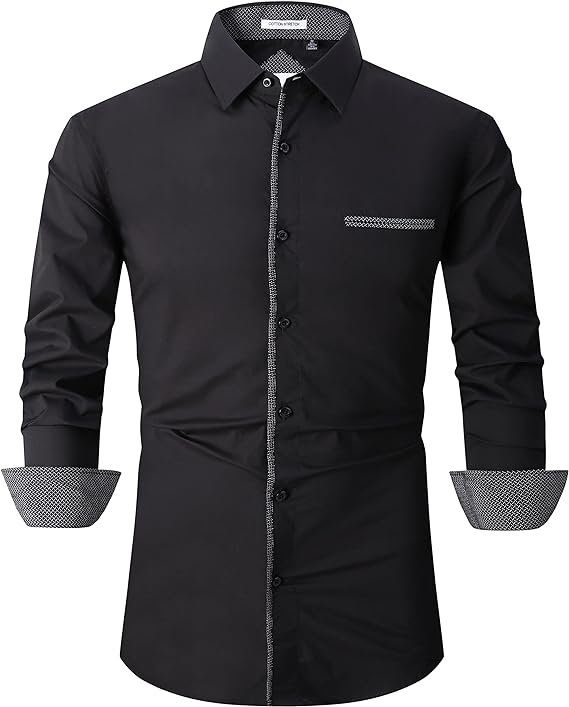} \\
            \hline
            \includegraphics[width=0.1\textwidth]{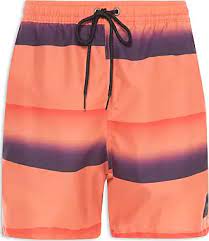} &
            \includegraphics[width=0.1\textwidth]{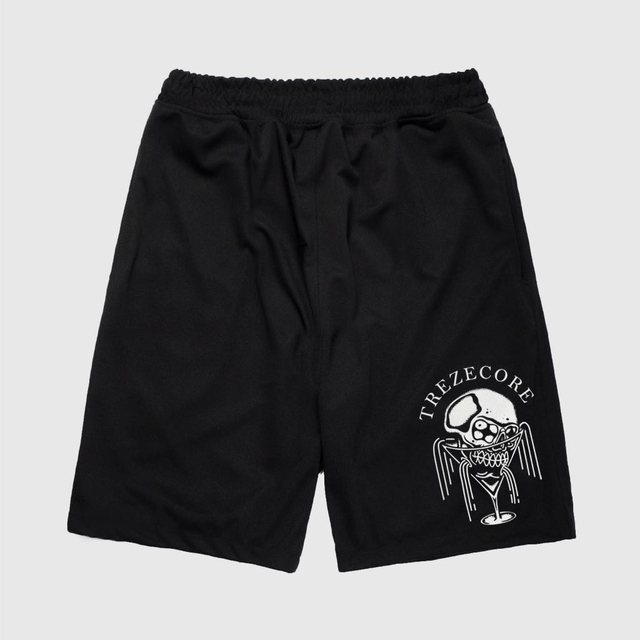} &
            \includegraphics[width=0.1\textwidth]{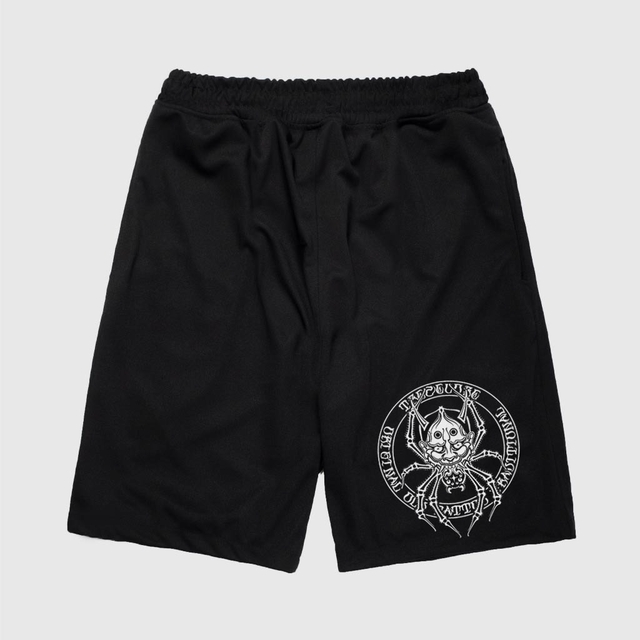} &
            \includegraphics[width=0.1\textwidth]{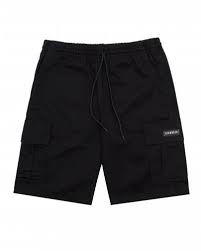} &
            \includegraphics[width=0.1\textwidth]{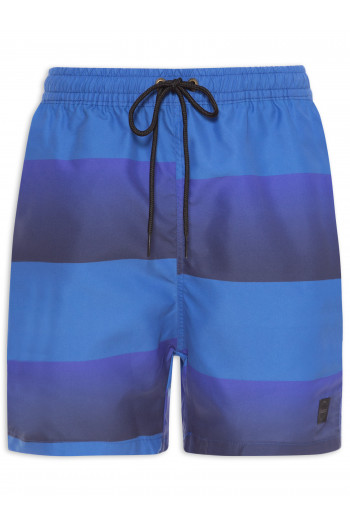} &
            \includegraphics[width=0.1\textwidth]{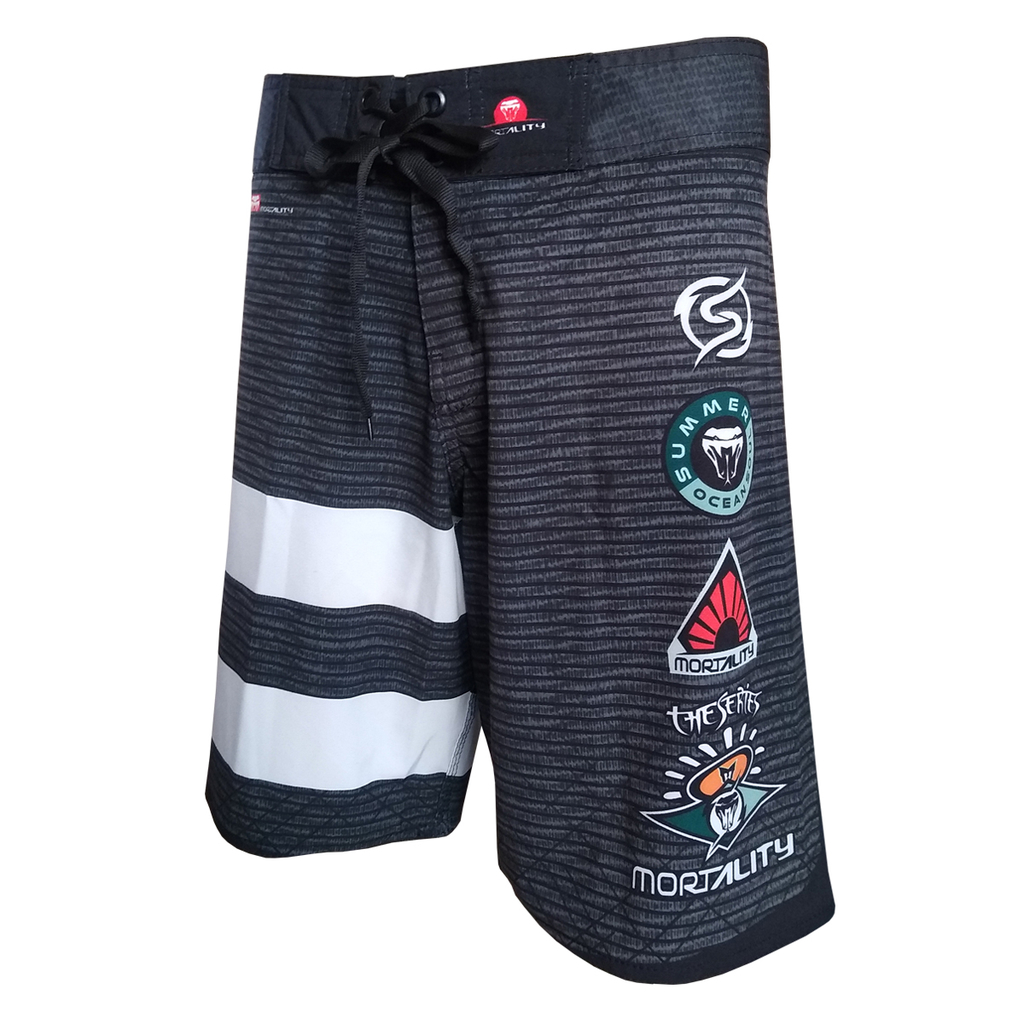} &
            \includegraphics[width=0.1\textwidth]{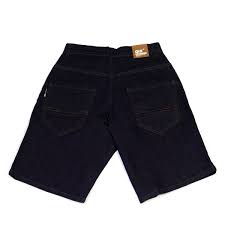} &
            \includegraphics[width=0.1\textwidth]{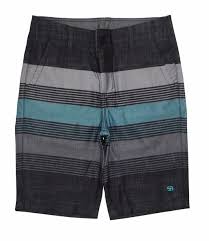} \\
            \hline
            \includegraphics[width=0.1\textwidth]{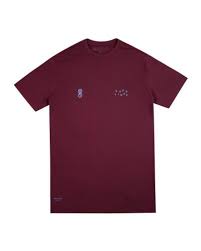} &
            \includegraphics[width=0.1\textwidth]{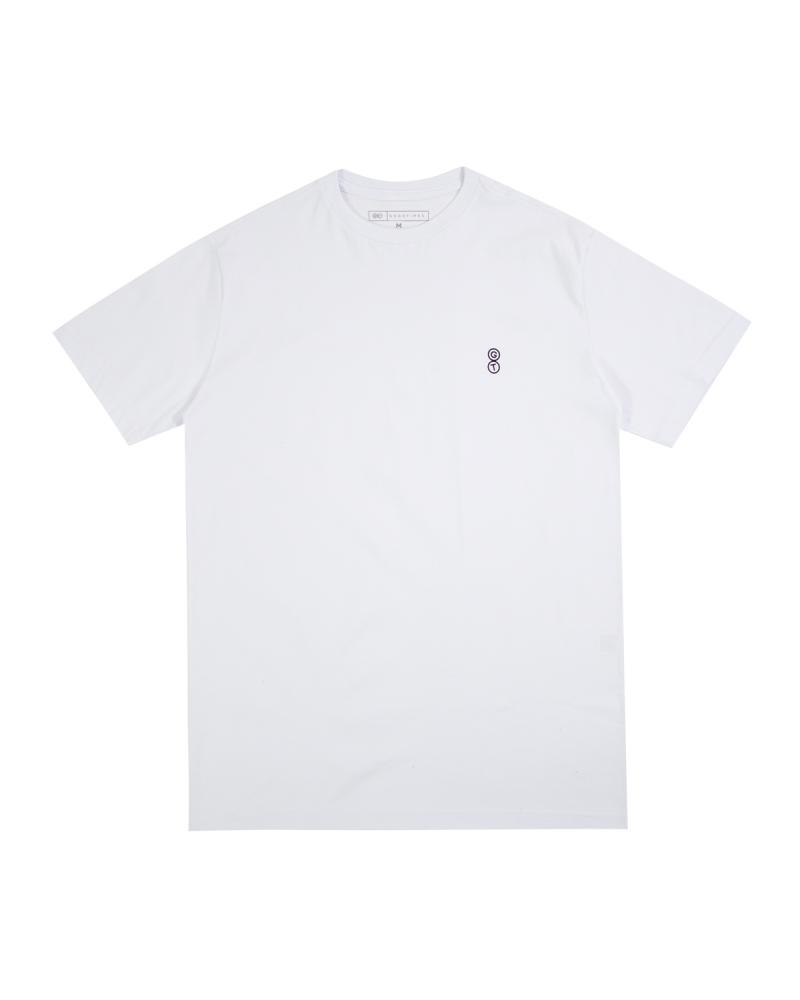} &
            \includegraphics[width=0.1\textwidth]{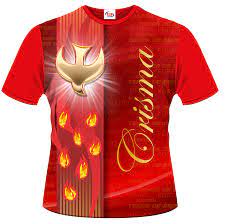} &
            \includegraphics[width=0.1\textwidth]{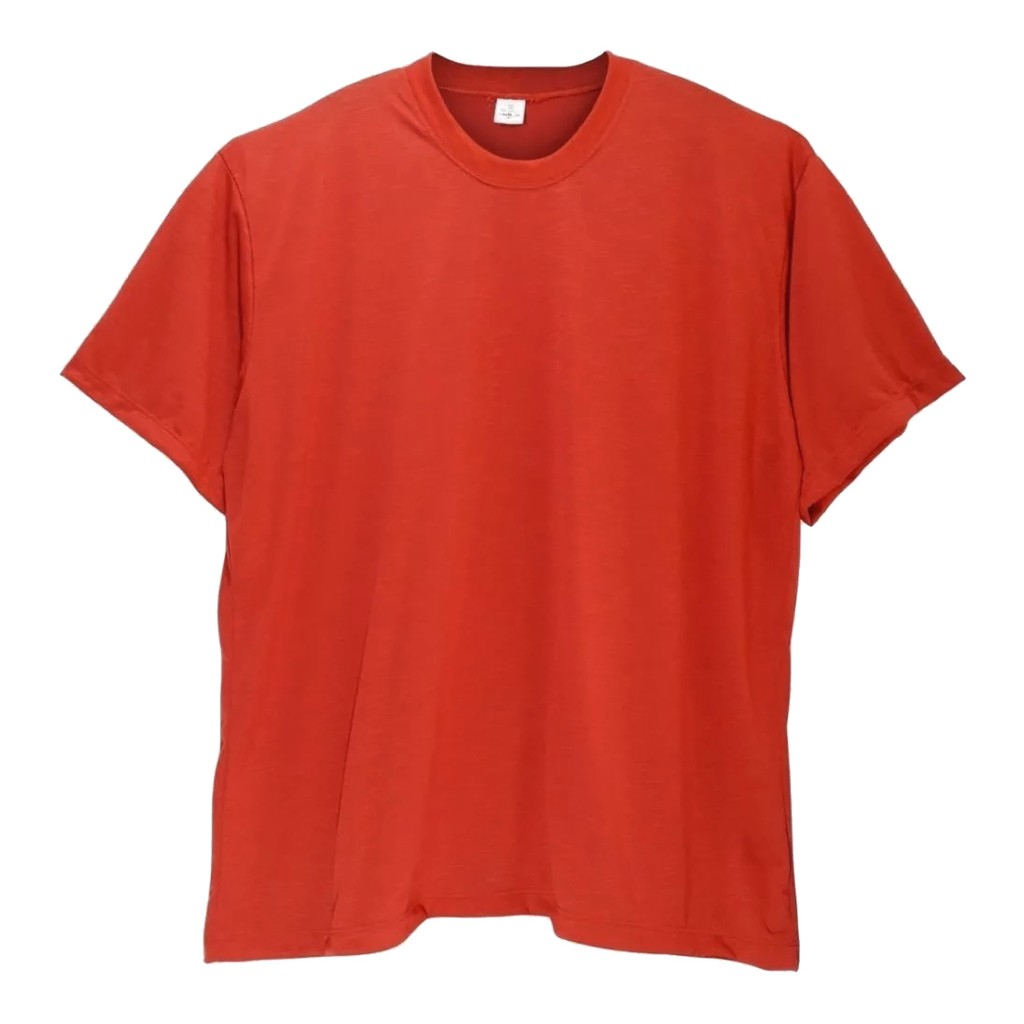} &
            \includegraphics[width=0.1\textwidth]{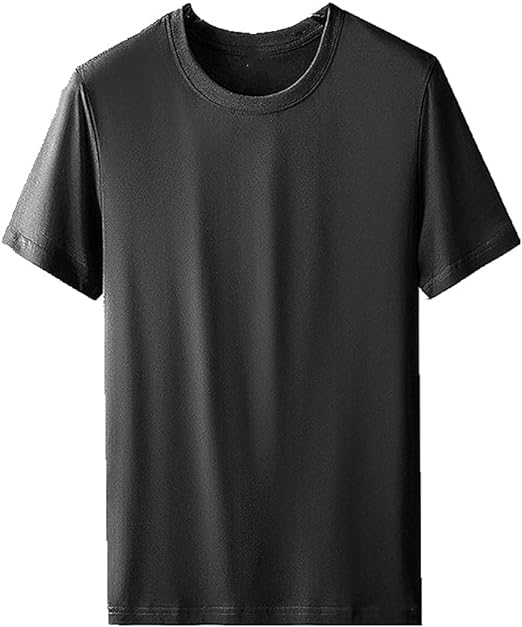} &
            \includegraphics[width=0.1\textwidth]{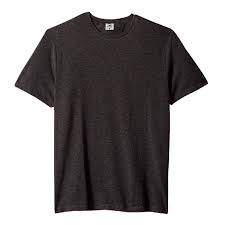} &
            \includegraphics[width=0.1\textwidth]{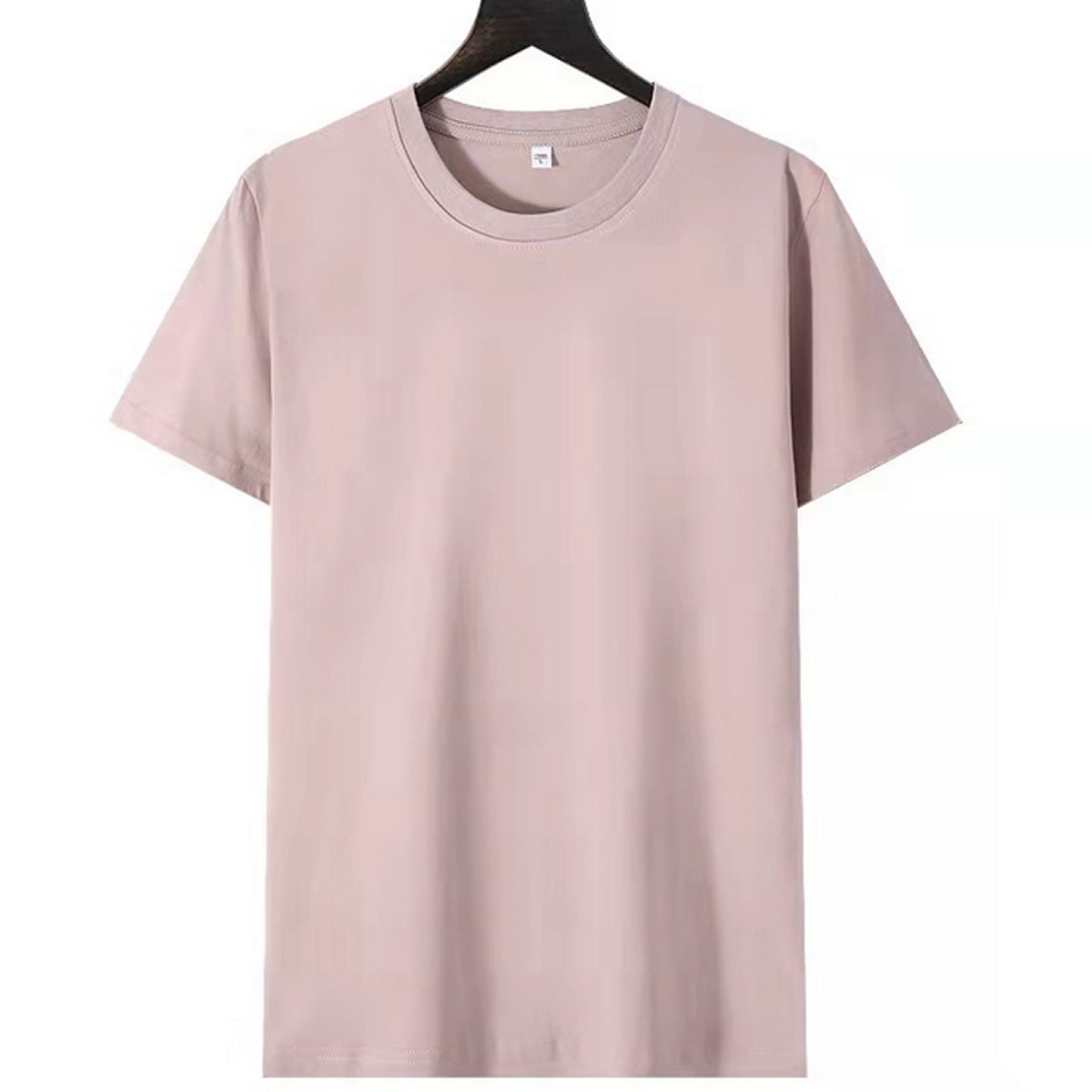} &
            \includegraphics[width=0.1\textwidth]{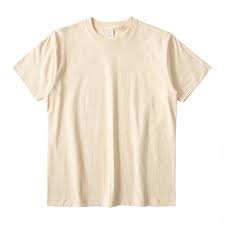} \\
            \hline
        \end{tabular}
    }
    \caption{Some images from four fashion categories used in our experiments. Each row corresponds to a different category.}
    \label{fig:imagelists-fashion-new}
\end{figure}

\subsubsection{Experiment 1}\label{exp:Approach-I-1}
\begin{figure}[H]
    \centering
    \includegraphics[width=\columnwidth]{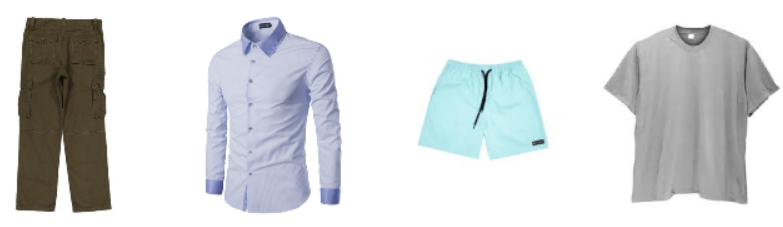}
    \caption{Representative images from (sub)categories whose instances appear in Fig.~\ref{fig:imagelists-fashion-new}.}
    \label{fig:reps-fashion-GAT}
\end{figure}

\begin{figure}[H]
    \centering
    \includegraphics[width=0.5\linewidth]{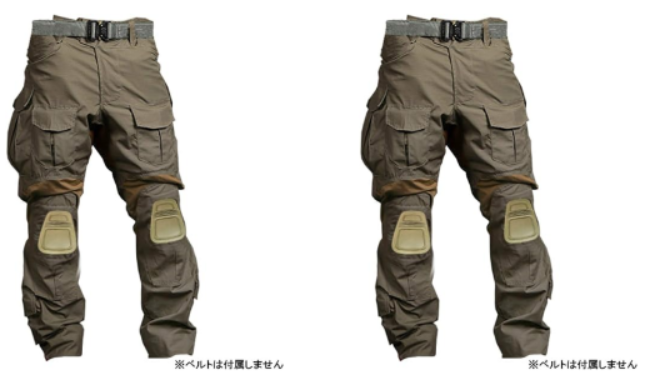}
    \caption{Query image and its best match in its identified category.}
    \label{fig:best_match_I-1}
\end{figure}

A step-by-step description of our GAT-AE based algorithmic framework is listed below:
\begin{itemize}
    \item We extract a 512-dimensional feature vector for each image using the pre-trained CLIP (ViT-B\slash 32)
 model.
    \item We stack the feature vectors into a tensor, where each row corresponds to a feature vector representing an image. We use this feature tensor as an input for the GAT-AE.
    \item The autoencoder's encoder comprises two GATConv layers followed by a latent layer that maps the data to a $256$-dimensional space. The decoder consists of fully connected layers that reconstruct the original features from the latent representation.
    \item  We construct a k-NN graph where each image node is connected to its k most similar images based on feature similarity. This graph structure enables the model to learn contextual relationships among images.
    \item After training, we compute the latent representation for each image and then average the latent vectors within each folder (i.e. listing) to obtain a representative model $\mathcal{R}_i$ for the image listing $\mathcal{L}_i$.
    \item During inference, we pass the query image through the same pipeline to obtain its context-aware latent as described in Section~\ref{sec:representative_model_I}. Then, we compare this model with $\{\mathcal{R}_i\}$ to determine the category of the query image.
  \item  Finally, we compute cosine similarity between the query image's latent vector and each listing’s representative. The listing with the highest similarity score is selected as the best match for the query image. 
    \end{itemize}
\noindent Here, the graph structure defines a fully connected graph where each image is a node connected to every other image via edges. We use the Adam optimizer to update the model’s weights and mean squared error (MSE) to measure the reconstruction loss of the features. We use ReLU as the activation function throughout the network. After training, we compute the cosine similarity between the latent representations, and define the average (i.e. the centroid) latent vector of each listing as its representative model. This representative is then used for comparison with query images during classification. Figure~\ref{fig:reps-fashion-GAT} illustrates the representatives of the categories, and Figure~\ref{fig:best_match_I-1} illustrates the query image with its match in its correctly identified category.


\begin{table}[!t]
\centering
\renewcommand{\arraystretch}{1.1}
\setlength{\tabcolsep}{2pt} 
\caption{Summary of training and inference logs for the GAT-AE based image retrieval pipeline.}
\label{tab:gat-log-summary}
\begin{tabular}{@{}p{0.35\columnwidth} p{0.55\columnwidth}@{}}
\toprule
\textbf{Steps} & \textbf{Output} \\
\midrule
Loaded images & 160 images from folders \\
KNN graph & Built with 1600 edges \\
Training & GAT autoencoder... \\
Epoch 2 & Loss: 0.218270 \\
Epoch 50 & Loss: 0.030119 \\
Epoch 100 & Loss: 0.019056 \\
Epoch 150 & Loss: 0.014004 \\
Epoch 200 & Loss: 0.010991 \\
Representatives & Saved representative vectors for each folder \\
\midrule
Similarity to (pants) & 0.7324 \\
Similarity to (shirt) & 0.6007 \\
Similarity to (shorts) & 0.5844 \\
Similarity to (t-shirt) & 0.6386 \\
Predicted category & \texttt{pants} \\
Closest image & \texttt{pants/01.jpg} \\
Closest similarity & 1.0000 \\
\bottomrule
\end{tabular}
\end{table}

\vspace{-1ex}

\subsubsection{Experiment 2}\label{exp:Approach-I-2}
Below, we give a brief review of the algorithm we used to conduct the same experiment using the \textit{histogram of oriented gradients} (HOG) method.
\begin{enumerate}[nosep]
    \item Representative identification:\\
 We first extract features from the images in a listing using the \textit{hog} method from the scikit-image library's \textit{feature} attribute. Then, we derive the feature vectors for all the images and determine the image that is the most similar to all the other images as the representative of the listing. 
\item Finding the category of a given image:\\
We compare the representative of the query image to these representative images to determine its subcategory. 
\end{enumerate}
\begin{figure}[H]
    \centering
    \includegraphics[width=\linewidth]{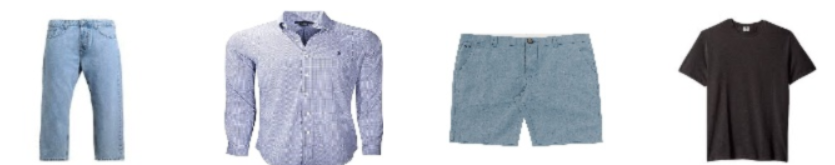}
    \caption{Representative images from categories whose instances appear in Figure~\ref{fig:imagelists-fashion-new}.}
    \label{fig:reps-fashion-HOG}
\end{figure}
\noindent Both GAT-AE based approach and the HOG-based approach yield the same best match for the given query image despite identifying category representatives differently (See Figure~\ref{fig:reps-fashion-HOG} and Figure~\ref{fig:reps-fashion-GAT}). The predicted category accuracy is observed as $0.7372$ since the query’s feature vector was most similar to the \textit{t-shirt} representative with a cosine similarity score of $0.7372$.

\begin{figure}[H]
    \centering
    \renewcommand{\arraystretch}{1.2}
    \resizebox{\columnwidth}{!}{
        \begin{tabular}{|c|c|c|}
            \hline
            \includegraphics[width=0.3\textwidth]{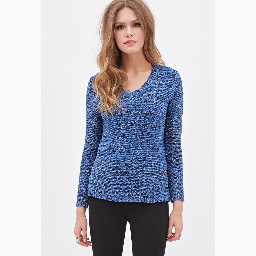} &
            \includegraphics[width=0.3\textwidth]{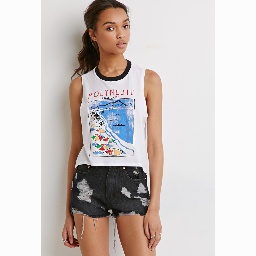} &
            \includegraphics[width=0.3\textwidth]{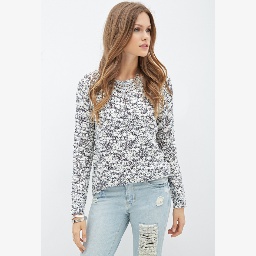} \\
            \hline
        \end{tabular}
    }
    \caption{Representative images for the listings shown in Fig.~\ref{fig:imagelists}.}
    \label{fig:reps}
\end{figure}

\begin{figure}[H]
    \centering
    \renewcommand{\arraystretch}{1.2}
    \resizebox{\columnwidth}{!}{
        \begin{tabular}{|c|c|c|c|c|c|c|c|}
            \hline
            \includegraphics[width=0.1\textwidth]{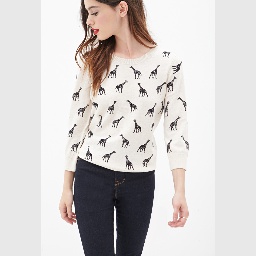} &
            \includegraphics[width=0.1\textwidth]{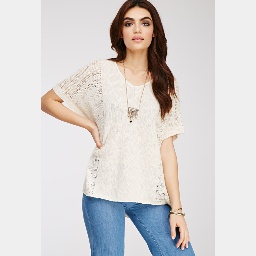} &
            \includegraphics[width=0.1\textwidth]{CC1/3.jpg} &
            \includegraphics[width=0.1\textwidth]{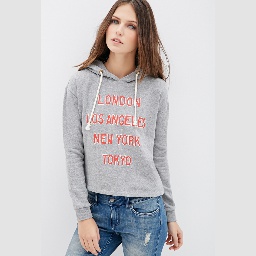} &
            \includegraphics[width=0.1\textwidth]{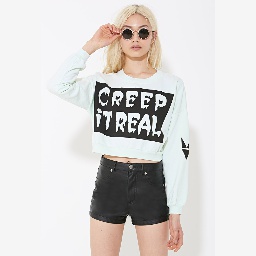} &
            \includegraphics[width=0.1\textwidth]{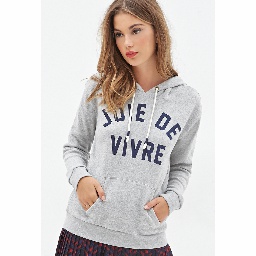} &
            \includegraphics[width=0.1\textwidth]{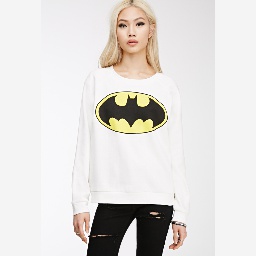} &
            \includegraphics[width=0.1\textwidth]{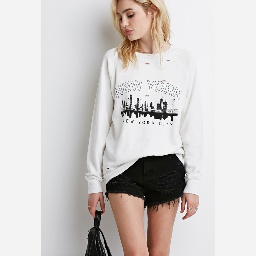} \\
            \hline
            \includegraphics[width=0.1\textwidth]{CC2/1.jpg} &
            \includegraphics[width=0.1\textwidth]{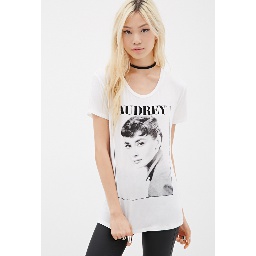} &
            \includegraphics[width=0.1\textwidth]{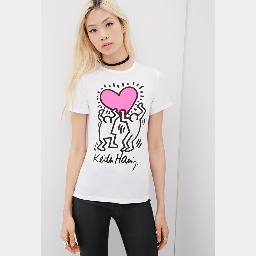} &
            \includegraphics[width=0.1\textwidth]{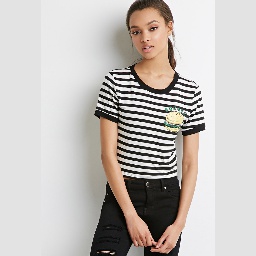} &
            \includegraphics[width=0.1\textwidth]{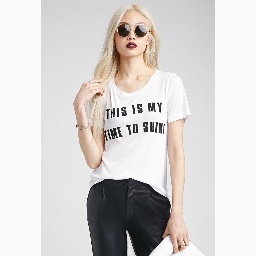} &
            \includegraphics[width=0.1\textwidth]{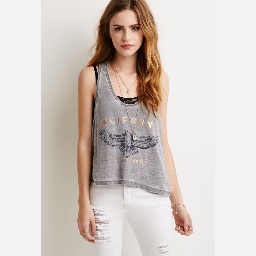} &
            \includegraphics[width=0.1\textwidth]{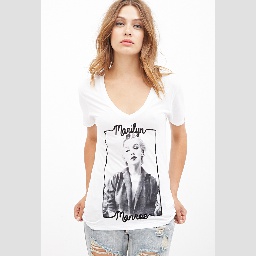} &
            \includegraphics[width=0.1\textwidth]{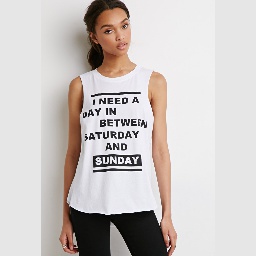} \\
            \hline
            \includegraphics[width=0.1\textwidth]{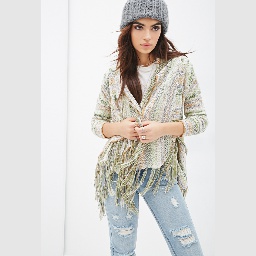} &
            \includegraphics[width=0.1\textwidth]{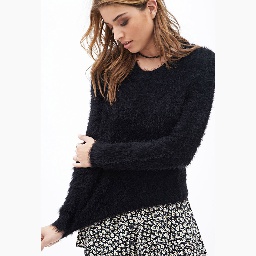} &
            \includegraphics[width=0.1\textwidth]{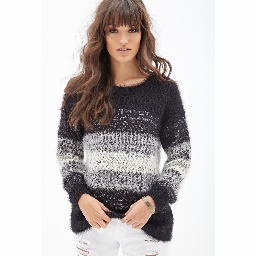} &
            \includegraphics[width=0.1\textwidth]{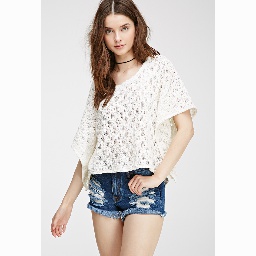} &
            \includegraphics[width=0.1\textwidth]{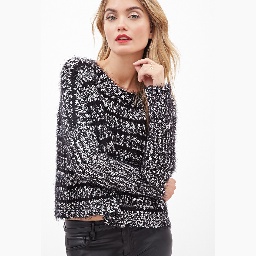} &
            \includegraphics[width=0.1\textwidth]{CC3/6.jpg} &
            \includegraphics[width=0.1\textwidth]{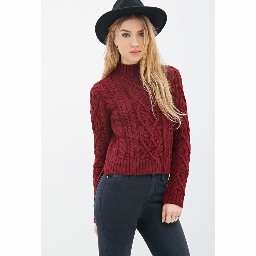} &
            \includegraphics[width=0.1\textwidth]{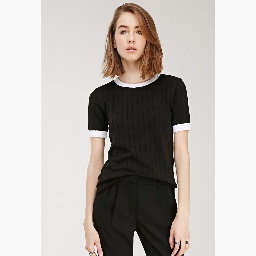} \\
            \hline
        \end{tabular}
    }
    \caption{Each row exhibits the images in a listing.}
    \label{fig:imagelists}
\end{figure}

\noindent Note that this method can also be used to categorize listings in a given directory, that is, if two listing representatives are sufficiently similar we may assume that the listings they represent belong to the same category. Figure~\ref{fig:reps} shows the representative images identified for the listings forming the rows of Figure~\ref{fig:imagelists} by using the \textit{hog} method. 

\noindent Then, the similarity matrix between these representative images is as follows:
\[
\begin{bmatrix}
1.0 & 0.67333986 & 0.81033915 \\
0.67333986 & 1.0 & 0.69574974 \\
0.81033915 & 0.69574974 & 1.0
\end{bmatrix}
\]
\noindent Thus, the most similar representative images in Figure~\ref{fig:reps} are identified as the images on the left and on the right with a
similarity score of $0.8103391471657959$. If the similarity threshold defined for category identification is lower than this score, we can merge the first and third listings claiming that they belong to the same category. 
\subsubsection{Experiment 3}\label{exp:Approach-I-3}
\begin{figure}[H]
    \centering
    \includegraphics[width=\columnwidth]{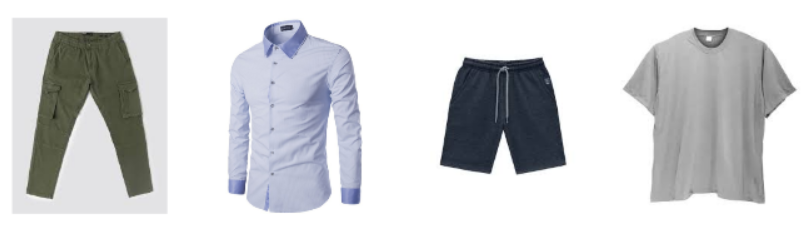}
    \caption{Representative images from categories whose instances appear in Fig.~\ref{fig:imagelists-fashion-new}.}
    \label{fig:reps-fashion-new}
\end{figure}

\begin{figure}[H]
    \centering
    \includegraphics[width=0.3\linewidth]{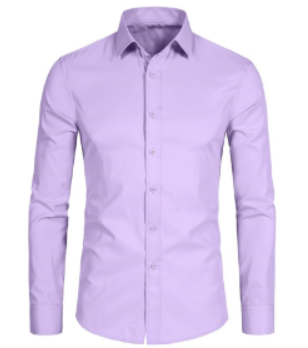}
    \caption{Query image.}
    \label{fig:query_image-fashion}
\end{figure}

A step-by-step description of our CLIP-based algorithmic framework is listed below:
\begin{itemize}
\item We use a pre-trained CLIP model (ViT-B\slash 32)
to extract a 512-dimensional feature vector for each image, and apply it to each image after resizing and normalization.
\item We stack all extracted feature vectors into a matrix, where each row corresponds to an image. We use this feature matrix in categorization and image retrieval.
\item For each category, we compute the average of the CLIP feature vectors of all images in that category, and identify the image most similar to it in within the category as the representative image of the category.
\item During inference, we pass a query image through the same feature extraction pipeline, and compare its feature vector with all category representatives to find the most similar category.
\item To retrieve the most similar image within that predicted category, we compute the cosine similarity between the query vector and all image vectors in the predicted category, selecting the one with the highest similarity.
\end{itemize}
Suppose we have the image listings in Figure~\ref{fig:imagelists-fashion-new}. Figure~\ref{fig:reps-fashion-new} shows the representative images identified for all the categories with instances illustrated along the rows. And we want to determine the listing of the image in Figure~\ref{fig:query_image-fashion}. 

\begin{figure}[H]
    \centering
    \includegraphics[width=0.6\linewidth]{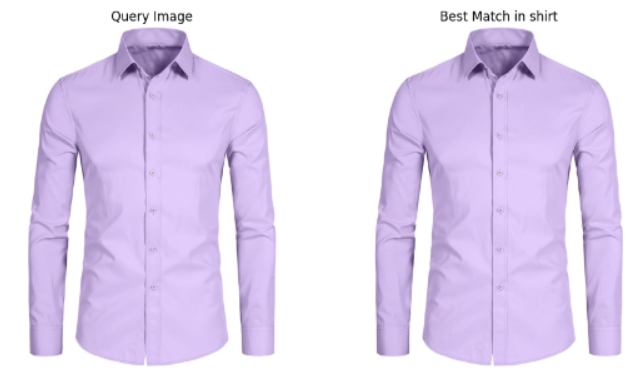}
    \caption{Query image and its best match in its identified category.}
    \label{fig:query_best_match_image-fashion}
\end{figure}

\noindent Our methodology determines the correct category, which in this case is the category labelled as \textit{t-shirt} (See Figure~\ref{fig:query_best_match_image-fashion}), and the similarity score for the best match in this category is calculated as $0.9330$. 

\begin{table}[!t]
\centering
\renewcommand{\arraystretch}{1.1}
\setlength{\tabcolsep}{2pt} 
\caption{Summary of classification results and query similarities.}
\label{tab:classification-summary}
\begin{tabular}{@{}p{0.38\columnwidth} p{0.56\columnwidth}@{}}
\toprule
\textbf{Step} & \textbf{Output} \\
\midrule
Feature extraction & Extracted features for 160 images \\
Prototypes & Computed category prototypes and representative images \\
\midrule
Categorization accuracy & 97\% \\
\midrule
\textbf{Category} & \textbf{Precision / Recall / F1-score} \\
pants & 0.91 / 1.00 / 0.95 \\
shirt & 0.98 / 1.00 / 0.99 \\
shorts & 1.00 / 0.90 / 0.95 \\
t-shirt & 1.00 / 0.97 / 0.99 \\
\midrule
Macro average & 0.97 / 0.97 / 0.97 \\
Weighted average & 0.97 / 0.97 / 0.97 \\
\midrule
\textbf{Query similarities} & -- \\
pants & 0.7860 \\
shirt & 0.9296 \\
shorts & 0.7678 \\
t-shirt & 0.8093 \\
\midrule
Predicted category & \texttt{shirt} \\
Best matching image & \texttt{shirt/12.jpg} \\
\bottomrule
\end{tabular}
\end{table}

\noindent We also used the adjacency matrix that we obtained to identify representatives. We computed the degree centrality of the images by summing the rows of the adjacency matrix. 
The image with the highest degree centrality is the most similar to all the other images in the relevant category since it is connected to the greatest number of images based on the cosine similarity threshold. We identified this central image as the category representative. As a result, we obtained the same category representatives and the same best match for the query image. 
\subsection{Applications of Approach II}\label{sec:apps-II}
\noindent As in Section~\ref{sec:apps-I}, we use the Adam optimizer to update the model's weights and
mean squared error (MSE) to measure the reconstruction loss of the features. We use ReLU as the activation function and we compute the cosine similarity as a similarity measure. \\
\indent In the experiments below, we apply our methodology to the \textit{Natural Images} dataset retrieved from \citep{kaggle_nature}. We utilize the categories consisting of images of \textit{vehicles}, namely, \textit{cars}, \textit{planes}, and \textit{motorbikes}. Thus, the main category is known to be \textit{vehicles}. In addition, we have three distinct subcategories rather than random listings belonging to this main category. We first identify the representatives of each (sub)category, and then determine the category of a query image. Finally, we find the most similar image to the query image within its identified category. For example, we first identify the query image as an image of a \textit{motorbike}, then we find the most similar motorbike avaliable in the dataset. \\

\begin{figure}[H]
    \centering
    \renewcommand{\arraystretch}{1.5}
    \resizebox{\columnwidth}{!}{
        \begin{tabular}{|c|c|c|c|c|c|c|c|}
            \hline
            \includegraphics[width=0.1\textwidth, height=0.15\textwidth]{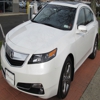} &
            \includegraphics[width=0.1\textwidth,height=0.15\textwidth]{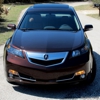} &
            \includegraphics[width=0.1\textwidth,height=0.15\textwidth]{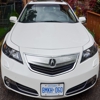} &
            \includegraphics[width=0.1\textwidth,height=0.15\textwidth]{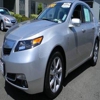} &
            \includegraphics[width=0.1\textwidth,height=0.15\textwidth]{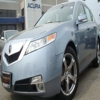} &
            \includegraphics[width=0.1\textwidth,height=0.15\textwidth]{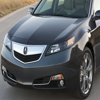} &
            \includegraphics[width=0.1\textwidth,height=0.15\textwidth]{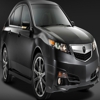} &
            \includegraphics[width=0.1\textwidth,height=0.15\textwidth]{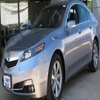} \\
            \hline
            \includegraphics[width=0.1\textwidth, height=0.15\textwidth]{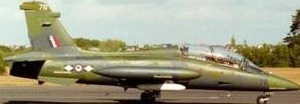} &
            \includegraphics[width=0.1\textwidth, height=0.15\textwidth]{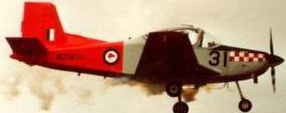} &
            \includegraphics[width=0.1\textwidth, height=0.15\textwidth]{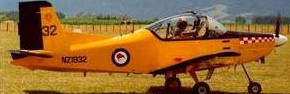} &
            \includegraphics[width=0.1\textwidth, height=0.15\textwidth]{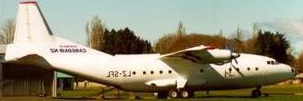} &
            \includegraphics[width=0.1\textwidth, height=0.15\textwidth]{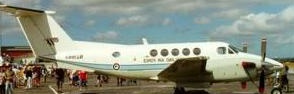} &
            \includegraphics[width=0.1\textwidth, height=0.15\textwidth]{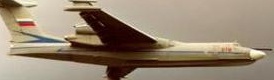} &
            \includegraphics[width=0.1\textwidth, height=0.15\textwidth]{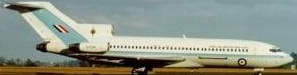} &
            \includegraphics[width=0.1\textwidth, height=0.15\textwidth]{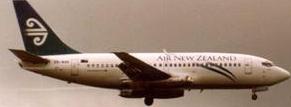} \\
            \hline
            \includegraphics[width=0.1\textwidth, height=0.15\textwidth]{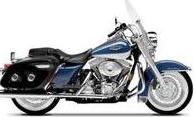} &
            \includegraphics[width=0.1\textwidth, height=0.15\textwidth]{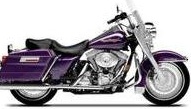} &
            \includegraphics[width=0.1\textwidth, height=0.15\textwidth]{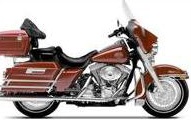} &
            \includegraphics[width=0.1\textwidth, height=0.15\textwidth]{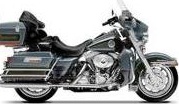} &
            \includegraphics[width=0.1\textwidth, height=0.15\textwidth]{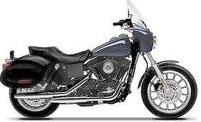} &
            \includegraphics[width=0.1\textwidth, height=0.15\textwidth]{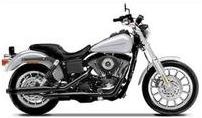} &
            \includegraphics[width=0.1\textwidth, height=0.15\textwidth]{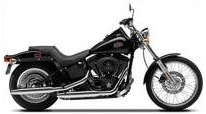} &
            \includegraphics[width=0.1\textwidth, height=0.15\textwidth]{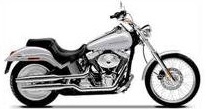} \\
            \hline
        \end{tabular}
    }
    \caption{Example images from three vehicle categories: cars, airplanes, and motorbikes. Each row represents a distinct category.}
    \label{fig:imagelists-vehicles}
\end{figure}

\subsubsection{Experiment 1}

   In this application, we generate each category's representative vector as the mean of the latent embeddings produced by the GAT-AE as described in Section~\ref{sec:approach_ii_rep_construct}.
 
\begin{figure}[H]
    \centering
    \includegraphics[width=0.4\linewidth]{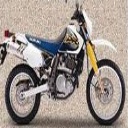}
    \caption{Query image}
    \label{fig:query_image-vehicle}
\end{figure}

\begin{figure}[H]
    \centering
    \renewcommand{\arraystretch}{1.2}
    \resizebox{\columnwidth}{!}{
        \begin{tabular}{|c|c|c|}
            \hline
            \includegraphics[width=0.3\textwidth]{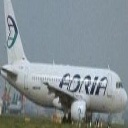} &
            \includegraphics[width=0.3\textwidth]{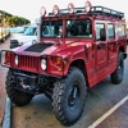} &
            \includegraphics[width=0.3\textwidth]{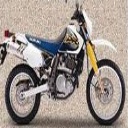} \\
            \hline
        \end{tabular}
    }
    \caption{Representative images corresponding to the listings shown in Fig.~\ref{fig:imagelists-vehicles}.}
    \label{fig:reps-HOG-vehicles}
\end{figure}

\begin{figure}[H]
    \centering
    \includegraphics[width=0.4\linewidth]{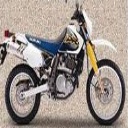}
    \caption{Most similar image in the category}
    \label{fig:most-similar-motorbike}
\end{figure}

 \noindent Suppose we have the image listings with instances illustrated in Figure~\ref{fig:imagelists-vehicles} and we want to find the category of the image in Figure~\ref{fig:query_image-vehicle}.
\noindent We obtain the representatives shown in See Figure~\ref{fig:reps-HOG-vehicles} for each category. 
\noindent The category of the query image is determined as the one representing the second row of images shown in Figure~\ref{fig:imagelists-vehicles}, that is the \textit{motorbike}, as a consequence of the results listed in Table~\ref{tab:similarity_scores-vehicles}. \noindent The most similar image to the query image in this category is the motorbike shown in Figure~\ref{fig:most-similar-motorbike} with a similarity score of 
$0.9881$.
\begin{table}[H]
\centering
\begin{tabular}{|l|c|}
\hline
\textbf{Category} & \textbf{Similarity score for the query image} \\
\hline
car & 0.5695 \\
motorbike & 0.9008 \\
airplane & 0.5412 \\
\hline
\end{tabular}
\caption{Similarity scores for different image categories}
\label{tab:similarity_scores-vehicles}
\end{table}

\begin{figure}[H]
    \centering
    \includegraphics[width=0.5\linewidth]{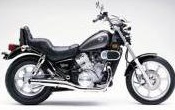}
    \caption{Query image}
    \label{fig:query_image-vehicle-2}
\end{figure}

\begin{figure}[H]
    \centering
    \includegraphics[width=0.5\linewidth]{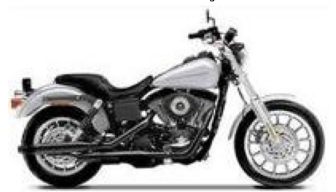}
    \caption{Most similar image in the category}
    \label{fig:most-similar-motorbike-2}
\end{figure}

\noindent When we do the search with an image not in the dataset which is shown in Figure~\ref{fig:query_image-vehicle-2}, we have the category determined correctly as before as a consequence of the similarity scores listed in Table~\ref{tab:similarity_scores-vehicles-2}. We retrieve the image in Figure~\ref{fig:most-similar-motorbike-2} as the most similar image with a similarity score of $0.9923$.

\begin{table}[H]
\centering
\begin{tabular}{|l|c|}
\hline
\textbf{Category} & \textbf{Similarity score for the query image} \\
\hline
car & 0.6166 \\
motorbike & 0.9474 \\
airplane & 0.6559 \\
\hline
\end{tabular}
\caption{Similarity scores for different image categories}
\label{tab:similarity_scores-vehicles-2}
\end{table}

\noindent Note that even if the retrieved image is identical to the query image, the context in the graph (i.e. the neighbors and position) and mathematical precision involved may lead to slightly different embeddings from a GAT. Although we're using the same image and the same GAT model, the structure of the graph during inference is different from when the image was embedded during training. This changes the neighbors used in the GAT framework, so the final latent embedding is not exactly the same.
Recall that GAT is not a pure function of the input features. Unlike a CNN, it depends on the input feature of a node, the features of its neighbors,the graph edges, and the learned attention weights.
Thus,  if we embed an image during training, it has a certain set of neighbors, whereas if we embed it during inference, we append it as a new node at the end and give it a different neighborhood. This yields a slightly different embedding, even for the same image.
Moreover, even if the image were in the same place in the graph, we would still get minor differences due to precision limits or layer-wise operations. 
As a result, we may observe that the cosine similarity drops from $1.0$ to $0.98$ or so.

\subsubsection{Experiment 2}
\begin{figure}[H]
    \centering
    \includegraphics[width=0.35\linewidth]{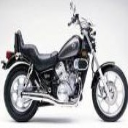}
    \caption{Query image}
    \label{fig:query_image_II-2}
\end{figure}

Now, we conduct the same experiment using the HOG method as in Section~\ref{sec:apps-I}. Suppose we have the image listings with instances illustrated in Figure~\ref{fig:imagelists-vehicles} and we want to find the category of the image in Figure~\ref{fig:query_image_II-2}.

\begin{figure}[H]
    \centering
    \includegraphics[width=0.35\linewidth]{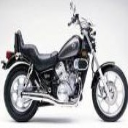}
    \caption{Most similar image to the query image in Figure~\ref{fig:query_image_II-2}}
    \label{fig:most-similar-motorbike-4}
\end{figure}

\noindent We identify the correct category with an accuracy quantified in terms of cosine similarity equal to $0.7196$, and retrieve the image we expected as the most similar(See Figure~\ref{fig:most-similar-motorbike-4}) with a similarity score of $1.0000$.

\subsubsection{Experiment 3}

\begin{figure}[H]
    \centering
    \includegraphics[width=\linewidth]{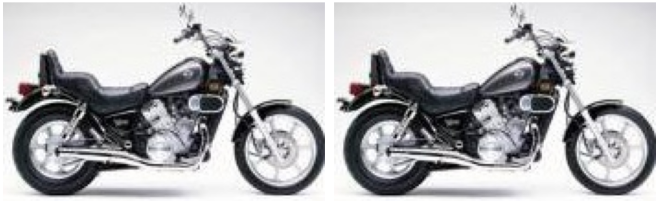}
    \caption{Query image on the left, most similar image on the right}
    \label{fig:most-similar-motorbike-3}
\end{figure}

We conduct the same experiment, using a pre-trained ResNet model instead of the HOG method or the GAT-AE. As can be seen in Table~\ref{tab:similarity_scores-vehicles-3}, we have the category of the query image determined correctly as in the previous experiments. We also retrieve the image in Figure~\ref{fig:most-similar-motorbike-3} as the most similar image with a similarity score of $1.0000$ as expected since the query image was selected from the listings analyzed.
\begin{table}[H]
\centering
\begin{tabular}{|l|c|}
\hline
\textbf{Category} & \textbf{Similarity score for the query image} \\
\hline
car & 0.7465 \\
motorbike & 0.9389 \\
airplane & 0.7204 \\
\hline
\end{tabular}
\caption{Similarity scores for different image categories}
\label{tab:similarity_scores-vehicles-3}
\end{table}
\noindent 
\subsection{Comments on Experimental Results}
\noindent There are some key differences between the graph-based approach and the centroid-based method employed in deriving the representatives of images.
The graph-based method and the centroid-based method differ in how they represent categories and determine image similarity. The centroid-based method summarizes each folder using the mean of its image feature vectors, treating the average as a proxy for the entire category. In contrast, the graph-based method models each folder as a similarity graph, selects the most central image based on its connections within that graph, and uses this image as the folder’s representative. After predicting the folder for a query image, the graph-based approach further refines its result by identifying the single most similar image within that folder using cosine similarity. This results in a more focused and interpretable match. While the centroid-based method is faster and simpler, the graph-based method provides a more structure-aware and precise representation by considering internal relationships between images in each category.\\
\indent These methods can be utilized in different case-scenarios. The centroid-based method is best suited for scenarios where execution time, simplicity, and scalability are important, such as in the case of large-scale image classification, initial filtering, or when category boundaries are well-defined and consistent. It works well when each folder can be reliably represented by a statistical average of its contents. On the other hand, the graph-based method is more appropriate when interpretability and intra-folder structure matter, such as in recommendation systems, prototype selection, or fine-grained similarity search. By capturing how individual images relate to one another within a folder, the graph-based approach can provide more meaningful representatives and finer control over similarity-based decisions, 
\section{Conclusion}\label{sec:conclusion}
In this manuscript, we presented a methodology for comparing and categorizing images that uses a graph neural network–based framework with CLIP and ResNet features.
Our methodology employed a GAT auto-encoder (GAT-AE) to build representative models for image categories or listings, enabling categorization and image retrieval via representative comparisons. We demonstrated applications of our method in identifying the category of  various query images by comparing their representatives against the pre-categorized representative models.  We also showed how the representatives and categories can be determined by using more conventional techniques such as the HOG or direct comparisons based on ResNet or CLIP features. We observed that these methods provide different levels of comparison: GAT-AE based and HOG-based approach focus on the structural properties of the images, whereas the direct feature comparison based approach emphasize on higher-level attributes, such as color and coarse shape. In our experiments, we used some fashion and natural image datasets with complex similarity patterns.

\section{Code availability}
\noindent The code is available at the GitHub repository \\
\url{https://github.com/dys8/Image_Categorization_Search_GAT_Reps}.

\section*{Acknowledgment}\noindent This research was funded by Innovate UK as part of the project ``Multimodal Catalogue Search For Second Hand Apparel Valuations'' (10101553).


\end{document}